%% file: main.tex
\documentclass[lettersize,journal]{IEEEtran}
\usepackage{amsmath,amsfonts}
\usepackage{algorithmic}
\usepackage{array}
\usepackage[caption=false,font=normalsize,labelfont=sf,textfont=sf]{subfig}

\usepackage{textcomp}
\usepackage{stfloats}
\usepackage{url}
\usepackage{verbatim}
\usepackage{graphicx}

\usepackage{epsfig}
\usepackage{graphicx}
\usepackage{amsmath}
\usepackage{amssymb}

\usepackage{lipsum}
\usepackage{stfloats}
\usepackage{multicol}
\usepackage{multirow}
\usepackage{bm}
\usepackage{etoolbox}

\usepackage{tabulary}
\usepackage[table]{xcolor} 
\usepackage{paralist}
\usepackage{booktabs}
\usepackage{adjustbox}
\usepackage{xspace}
\usepackage{xcolor}
\definecolor{rblue}{rgb}{0,0.5,1}
\usepackage{hyperref}
\hypersetup{colorlinks=true,linkcolor=red,citecolor=rblue}

\def\eg{\textit{e.g.}} 
\def\ie{\textit{i.e}.}

\def\BibTeX{{\rm B\kern-.05em{\sc i\kern-.025em b}\kern-.08em
    T\kern-.1667em\lower.7ex\hbox{E}\kern-.125emX}}
\usepackage{balance}

\newcommand{\YKL}[1]{\textcolor{red}{#1}}

\newcommand{\JQ}[1]{\textcolor{blue}{#1}}

\begin{document}
\title{Minimalist and High-Quality Panoramic Imaging with PSF-aware Transformers}%
\author{Qi Jiang\IEEEauthorrefmark{1}, Shaohua Gao\IEEEauthorrefmark{1}, Yao Gao, Kailun Yang\IEEEauthorrefmark{2}, Zhonghua Yi, Hao Shi, Lei Sun, and Kaiwei Wang\IEEEauthorrefmark{2}%
\thanks{This work was supported in part by the National Natural Science Foundation of China (NSFC) under Grant No. 12174341 and in part by Hangzhou SurImage Technology Company Ltd.}%
\thanks{Q. Jiang, S. Gao, Y. Gao, Z. Yi, H. Shi, L. Sun, and K. Wang are with the State Key Laboratory of Modern Optical Instrumentation and the National Engineering Research Center of Optical Instrumentation, Zhejiang University, Hangzhou 310027, China.}%
\thanks{K. Yang is with the School of Robotics and the National Engineering Research Center of Robot Visual Perception and Control Technology, Hunan University, Changsha 410082, China.}%
\thanks{\IEEEauthorrefmark{1}Equal contribution.}%
\thanks{\IEEEauthorrefmark{2}Corresponding authors: Kaiwei Wang and Kailun Yang. (E-mail: wangkaiwei@zju.edu.cn, kailun.yang@hnu.edu.cn.)}%
}

\markboth{IEEE Transactions on Image Processing, July~2024}%
{Jiang \MakeLowercase{\textit{et al.}}: PALHQ}

\maketitle

\begin{abstract}
\input{Tex_content/abstract_revised}
\end{abstract}


\section{Introduction}
\input{Tex_content/Introduction_revised}

\input{Tex_content/contribution_revised}

\begin{figure}[!t]
  \centering
  \includegraphics[width=1.0\linewidth]{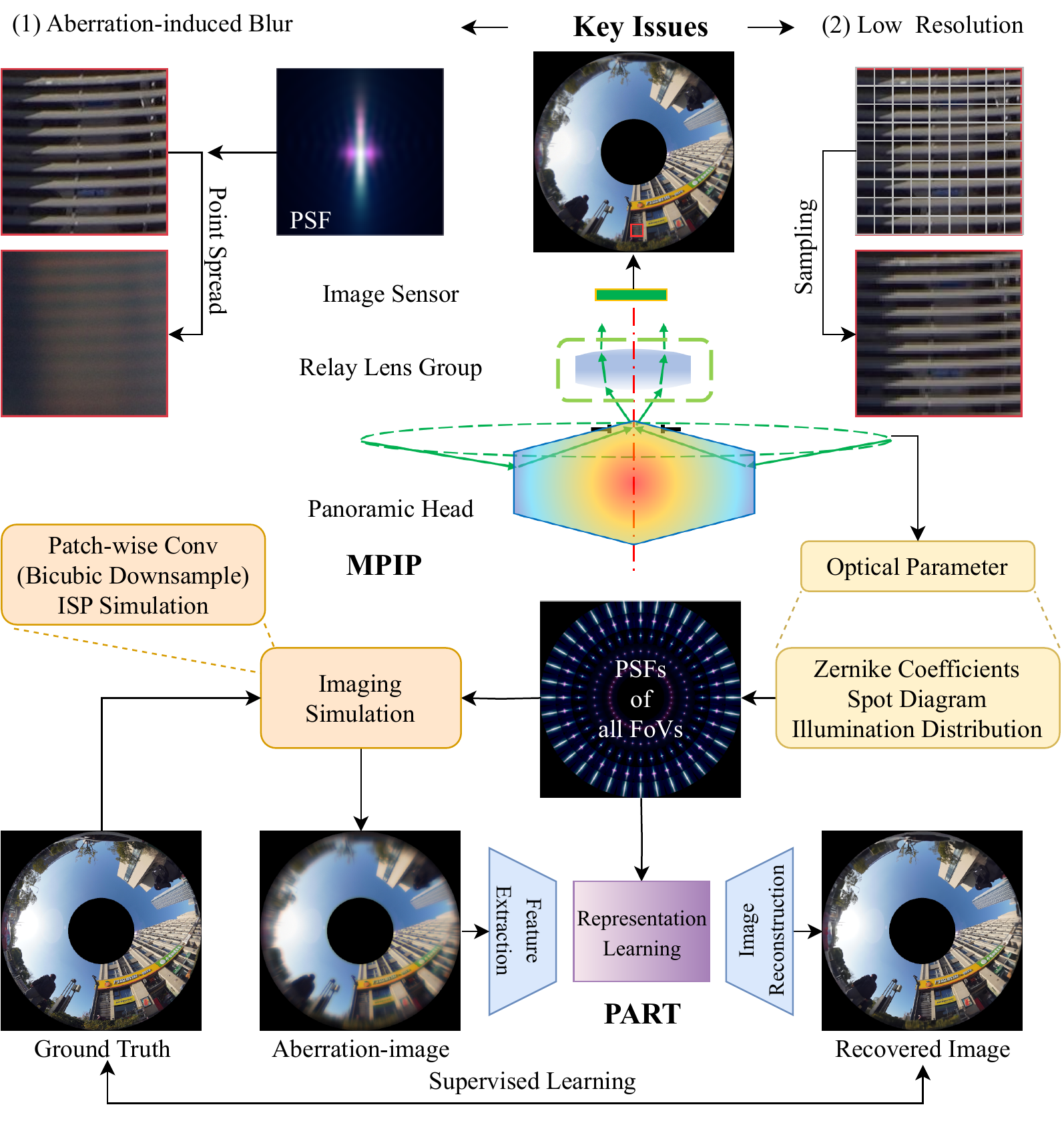}

  \caption{Overview of the proposed Panoramic Computational Imaging Engine (PCIE) for minimalist and high-quality panoramic imaging. To achieve the goal of panoramic imaging with a minimalist system, the number of optical components and the radius of MPIP are designed to be small, which brings two key issues of low image quality: (1) aberration-induced blur due to lack of enough lenses for aberration correction and (2) low resolution caused by limited image plane size. We introduce the PART, which is trained on synthetic data pairs generated by imaging simulation, to recover the low-quality aberration image with the guidance of PSF information. ``ISP'' denotes Image Signal Processing.}
  \label{fig:overview}

\end{figure}

\section{Related Work}
\input{Tex_content/related_work_revised}

\section{Minimalist Panoramic Imaging Prototype}
\label{sec:MPIP}
In this section, we set up a universal prototype for minimalist panoramic imaging systems based on modern panoramic lens designs (Sec.~\ref{sec:design}).
To address the issues induced by the reduced lens numbers and limited image plane size, two settings of tasks and benchmarks are defined in Sec.~\ref{sec:task} and Sec.~\ref{sec:benchmark}, respectively.
In Sec.~\ref{sec:simulation}, we describe the constructed imaging simulation model to generate synthetic image pairs for training learning-based methods. 

\subsection{Optical Design}
\label{sec:design}
To boost scene understanding with larger FoV, panoramic optical systems are emerging, including fisheye optical systems, refractive panoramic systems, panoramic annular optical systems, \textit{etc.}~\cite{gao2022review}.
In most modern designs of panoramic lenses~\cite{Cheng2016DesignOA, Zhang2020DesignOA, wang2022design}, a panoramic head is applied for collecting incident rays of 360{\textdegree} FoV, while a set of relay lenses is designed to bend the rays and correct the aberrations.
Based on the structure, we propose the Minimalist Panoramic Imaging Prototype (MPIP), including an essential panoramic head and a simple relay lens group, as shown in Fig.~\ref{fig:intro}(a). 

\begin{figure}[!t]
  \centering
  \includegraphics[width=0.8\linewidth]{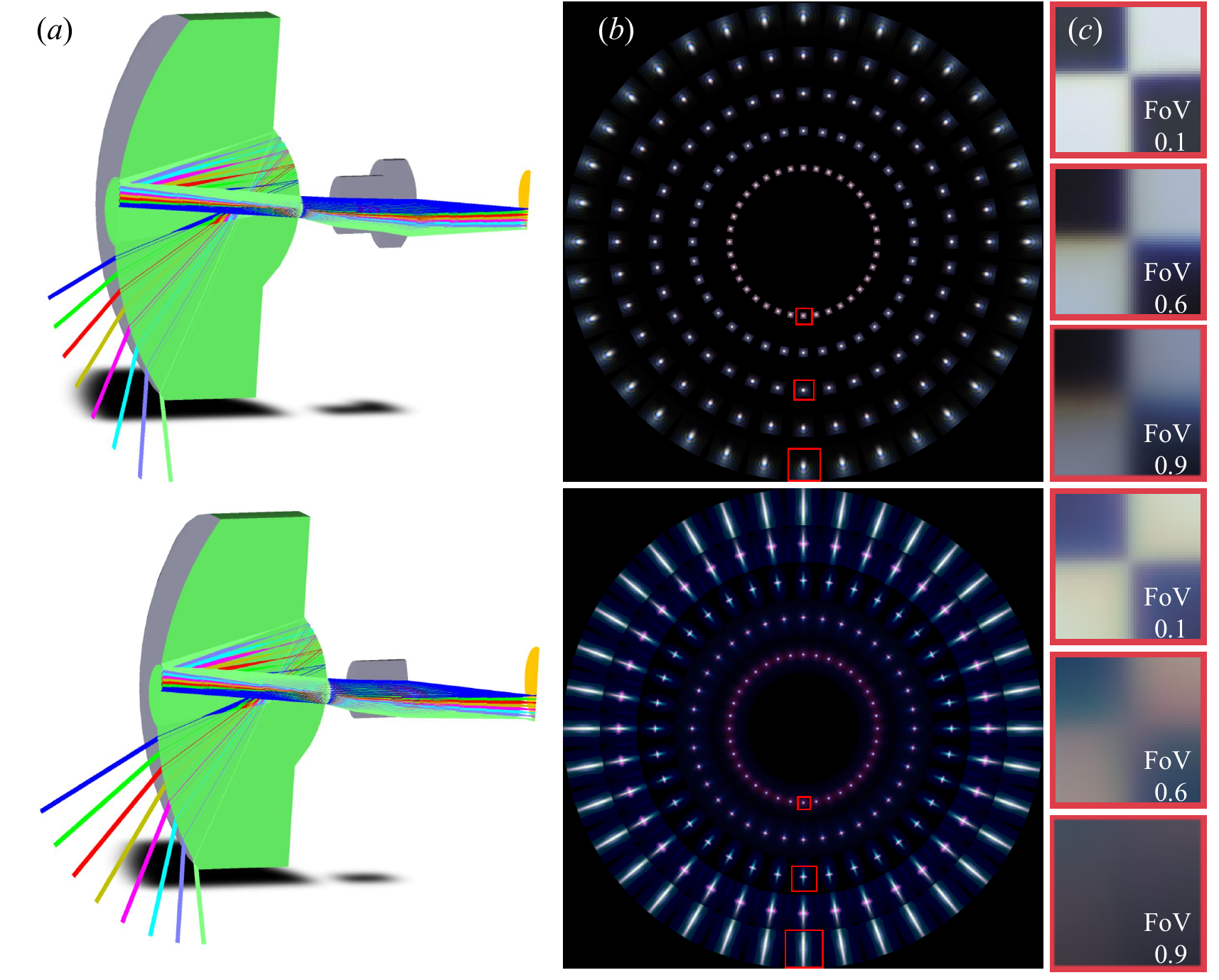}

  \caption{Two prototype samples of MPIP. Up: MPIP-P1, Down: MPIP-P2. (a) Optical path diagram. (b) Visualized PSF distributions. (c) The degraded checkerboard image patches of normalized FoVs $0.1$, $0.6$, and $0.9$ are captured by two MPIP samples. The minimalist optical design brings spatially-variant aberration-induced blur, especially for MPIP-P2 equipped with fewer lenses.}
  \label{fig:pal}

\end{figure}

Specifically, we adopt a more compact and efficient solution, \ie~Panoramic Annular Lens~(PAL)~\cite{greguss1986panoramic,powell1994panoramic}, in MPIP samples, where a catadioptric PAL head is equipped for 360{\textdegree} annular imaging. 
For minimalist design and convenient manufacture, spherical lenses are applied in the relay lens group and the number is reduced to fewer than $3$.
To illustrate the imaging results of different relay lens groups, we define MPIP-P1 and MPIP-P2 in Fig.~\ref{fig:pal}(a), whose relay lens group is composed of two lenses and a single lens, respectively.

The lack of enough lenses for aberration correction makes the imaging point spread from an ideal point, inducing spatially-variant PSFs with large kernel sizes, as shown in Fig.~\ref{fig:pal}(b). 
The average geometric spot radius of MPIP-P1 is $13.78 {\mu}m$, whereas that of MPIP-P2 is $46.26 {\mu}m$.
As a result, the captured images of MPIP suffer from spatially-variant aberration-induced blur, especially for MPIP-P2 with fewer lenses, as shown in Fig.~\ref{fig:pal}(c).

\subsection{Definition of Tasks}
\label{sec:task}

\begin{figure}[!t]
  \centering
  \includegraphics[width=1.0\linewidth]{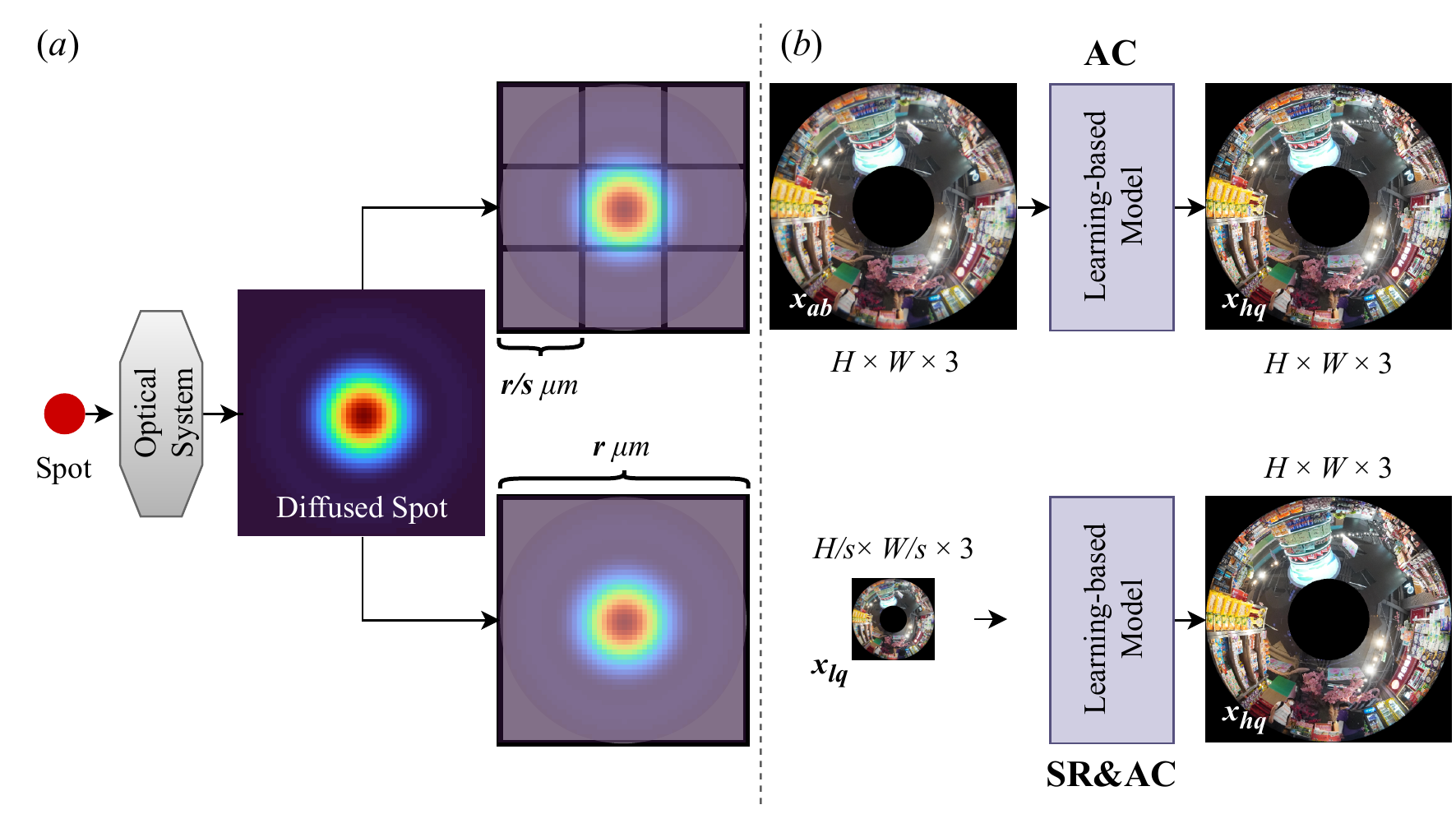}

  \caption{Illustration of two pipelines for processing aberration-images of MPIP. (a) Comparison of image sensors with different pixel sizes. For a diffused spot through the optical system with a fixed size, more pixels of the sensor with smaller pixel sizes and higher resolutions are affected. (b) Raised two tasks based on sensors with different pixel sizes: Aberration Correction (AC) and Super-Resolution and Aberration Correction (SR$\&$AC). In summary, we target the recovery of a high-quality image from an aberration image of MPIP.}
  \label{fig:task}

\end{figure}

In addition to the uncorrected optical aberrations, the limited image plane size due to the small aperture of MPIP presents the issue of image resolution.
To fit the small image plane of the MPIP, an image sensor with a smaller pixel size can be applied to maintain high resolution, but it makes the system more sensitive to aberration-induced blur. 
As shown in Fig~\ref{fig:task}(a), the diffused optical spot of fixed physical size affects more pixels for the sensor with smaller pixel sizes and higher resolution. 
The opposite solution with large pixel sizes is less sensitive to the diffused spot, but the reduced image resolution also brings degradation to the images, which is especially harmful to panoramic images with large FoV~\cite{yu2023osrt}. 

To address this dilemma, we propose two pipelines for solving the contradictory problems, as shown in Fig.~\ref{fig:task}(b), where a learning-based model is applied to process different image recovery tasks. 
For image sensors with smaller pixel sizes, we define the Aberration Correction (AC) task, where the goal is to recover a clear image ${x}_{hq}\in\mathbb{R}^{H{\times}W{\times}3}$ from a high-resolution input aberration-image ${x}_{ab}\in\mathbb{R}^{H{\times}W{\times}3}$.
Whereas for image sensor with larger pixel size, the Super-Resolution and Aberration Correction (SR$\&$AC) task is raised to recover a high-resolution aberration-free image ${x}_{hq}\in\mathbb{R}^{H{\times}W{\times}3}$ from a low-resolution input aberration-image ${x}_{{lq}}{\in}\mathbb{R}^{\frac{{H}}{{s}}\times \frac{{W}}{{s}}\times 3}$, where $s$ is the scale factor of SR.

\subsection{PALHQ: Established Dataset of High-Quality PAL Images}
\label{sec:benchmark}

The lack of high-quality image datasets for PAL comes as a bottleneck to the above tasks. 
A piece of previous work for CI of PAL, ACI~\cite{jiang2023annular}, unfolds the annular PAL images into perspective ones to utilize the publicly available datasets, \ie~DIV2K~\cite{Agustsson_2017_CVPR_Workshops}. 
However, the asymmetrical interpolation during unfolding induces extra image degradation, which further complicates the image degradation factors for MPIP. 
In addition, the annular image is more appealing for the simulation of aberrations in the original image plane and necessary for some vision tasks like PAL-based SLAM~\cite{chen2021panoramic_SLAM,wang2022lf}.
In the case of benchmarks for processing panoramic images, \eg, the ODI-SR dataset~\cite{deng2021lau} and the SUN360 panorama dataset~\cite{xiao2012recognizing}, which are taken via fisheye cameras, the imaging process is also quite different from that of PAL~\cite{gao2022review}.
These concerns raise an urgent request for high-quality panoramic annular image datasets.

To this intent, we propose PALHQ, a dataset of high-quality PAL images, to facilitate network training and evaluation of PAL-based low-level vision tasks. 
A well-designed PAL of $11$ lenses and a Sony $\alpha$6600 camera are applied to capture high-resolution PAL images with negligible primary aberrations. 
PALHQ contains $550$ clear PAL images with a resolution of $3152\times3152$, covering rich and varied scenes of indoor, natural, urban, campus, and scenic spots.
We divide PALHQ into $500$ images for the training set and $50$ images for the validation set (refer to the appendix for sample images of PALHQ).
In PCIE, we benchmark both AC and SR$\&$AC on PALHQ, where the corresponding aberration images are generated by the imaging simulation model depicted below.
Furthermore, PALHQ can be also transmitted to unfolded panoramas via equirectangular projection (ERP), which can support various panoramic image processing applications.

\subsection{Imaging Simulation Model}
\label{sec:simulation}
To quantitatively benchmark the raised two tasks and enable supervised training of learning-based models, paired aberration images and clear images are required. 
Following previous super-resolution works~\cite{zhang2021designing, wang2021real} and CI works~\cite{jiang2023annular, chen2022computational_mass}, we construct an imaging simulation model to generate synthetic aberration-images in batches.

The wave-based simulation pipeline with random perturbation in~\cite{jiang2023annular} is adopted to generate multiple aberration distributions directly on clear annular PAL images.
Specifically, the clear raw image $R$ is modulated by an optical system and then processed by ISP $\Gamma(\cdot)$ to produce the final imaging result $A$:

\begin{equation}
\label{eq:basic2}
A_{\theta}(x,y) = \Gamma[(\int{{r_{\lambda}}R_{\theta}(x,y)\otimes K_{\theta}(x,y,\lambda)d\lambda})\downarrow+N],
\end{equation}
where ${r_{\lambda}}$ is the wave response of the sensor. 
The noise~$N$ and the sampling process~$(\cdot)\downarrow$ of the image sensor are also included in the model.
We divide the image into patches for patch-wise convolution with PSFs $K_{\theta}(x,y,\lambda)$ under different FoV $\theta$. 
Different from~\cite{jiang2023annular}, the division of FoV is centrosymmetric for annular images as is shown in Fig.~\ref{fig:pal}(b). 
Through scalar diffraction integral~\cite{huggins2007introduction}, $K_{\theta}(x,y,\lambda)$ is calculated based on the wavefront $\Phi_{\theta}(x',y',\lambda)$ on exit pupil plane, which is described by Zernike polynomials~\cite{mahajan1994zernike} mathematically:
\begin{equation}
\label{eq:zernike}
\Phi_{\theta}(x',y',\lambda) = \sum_{n,m} {C^m_n}(\theta,\lambda){Z^m_n}(x',y'),
\end{equation}
where $C(\theta,\lambda)$ denotes Zernike coefficients under FoV $\theta$ and wavelength $\lambda$ and $Z$ refers to polynomials of the coordinate $(x',y')$ on exit pupil.
The combination of different $m$ and $n$ represents different orders.
Finally, we apply the random disturbance strategy in~\cite{jiang2023annular} to fine-tune the ideal $C(\theta,\lambda)$ from the $Zemax$ software, generating synthetic aberration images with diverse aberration distributions. 
 
\section{Low-Quality MPIP Images Recovery}
\label{sec:part}
\begin{figure*}[!t]
  \centering
  \includegraphics[width=0.9\linewidth]{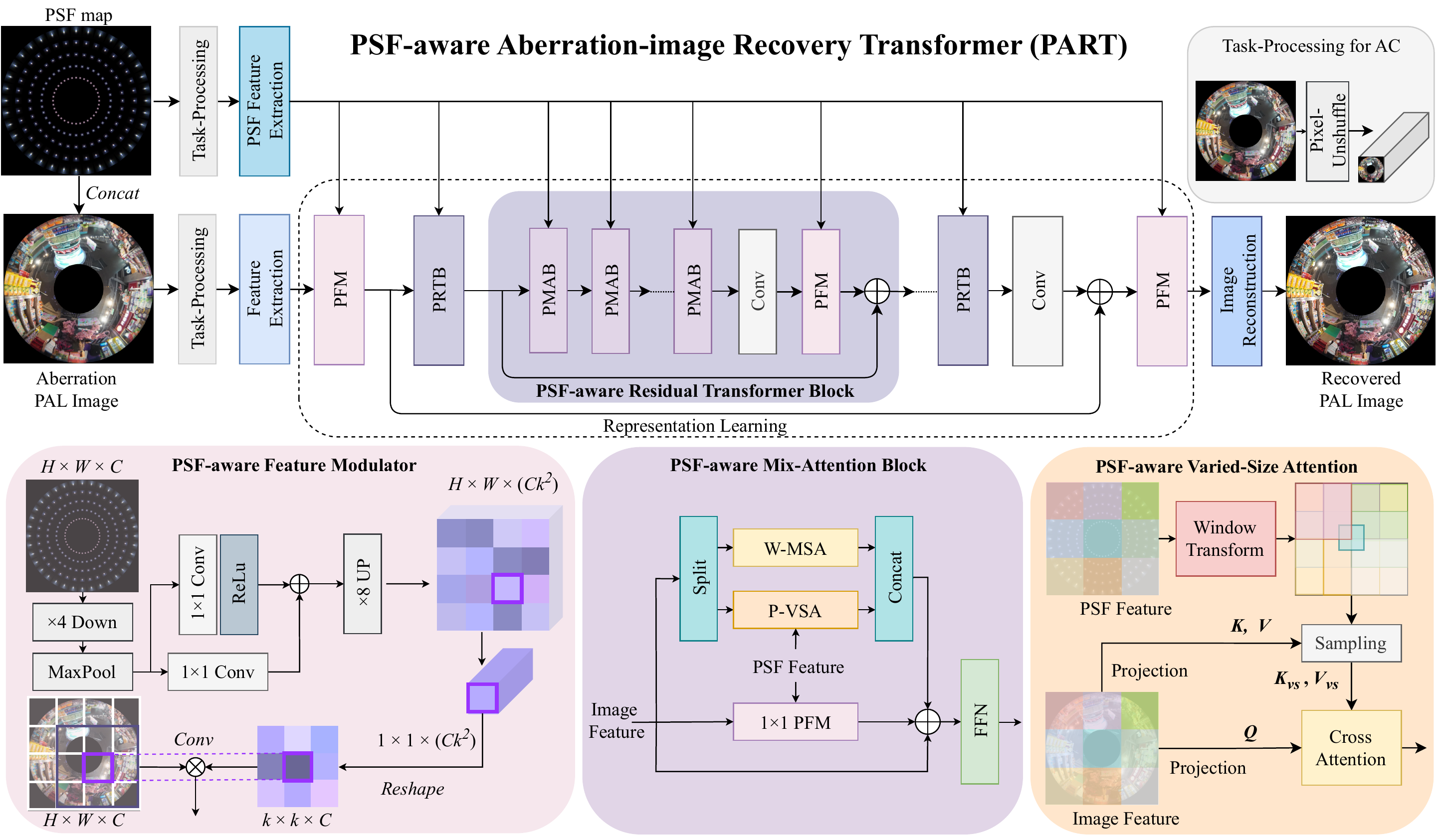}

  \caption{PART: Proposed PSF-aware Aberration-image Recovery Transformer. PART is established on a classical super-resolution paradigm~\cite{liang2021swinir,li2023efficient}, incorporating stages of feature extraction, representation learning, and image reconstruction, for dealing with both AC and SR$\&$AC. Task-Processing leverages the pixel-unshuffle operation~\cite{wang2021real} for AC to reduce the spatial size of high-resolution images, whereas no operation is entailed for SR$\&$AC. PSF-aware Residual Transformer Block (PRTB) is the basic block of representation learning, where we design PSF-aware Feature Modulator (PFM) and PSF-aware Mix-Attention Block (PMAB) to learn spatially-variant degradation features with the guidance of PSF features. The mixing of Window-based Multi-head Self-Attention (W-MSA), PSF-aware Varied-Size Attention (P-VSA), and PFM enable PMAB to capture both global and local dependencies adaptively. For an intuitive understanding of the PSF map, we visualize it via a form of PSF distributions in Fig.~\ref{fig:pal}.}
  \label{fig:part}

\end{figure*}

In this section, we describe the proposed learning-based model to recover low-quality MPIP images, as shown in Fig.~\ref{fig:part}.
The PSF information, characterizing the image degradation process, is represented as the PSF map, detailed in Sec.~\ref{sec:pi}, serving as one additional modality for our model.
With the PSF map, we design the PSF-aware Feature Modulator (PFM) and the PSF-aware Mix-Attention Block (PMAB), elaborated in Sec.~\ref{sec:pfm} and Sec.~\ref{sec:pmab}, respectively.
Then, the PSF-aware Aberration-image Recovery Transformer (PART) is established and introduced in Sec.~\ref{sec:network} as a transformer-based paradigm for the raised two tasks. 

\subsection{The Representation of PSF Information}
\label{sec:pi}
For non-blind optimization-based recovery methods in aberration correction, \eg~Wiener filter~\cite{wiener1949extrapolation}, PSFs $K$ of the system are exploited to predict the clear image $x_{hq}$ from the aberration-image $x_{ab}$ by deconvolution.
However, this method often fails when the PSFs deviate from the design stage during manufacture and require time-consuming strategies~\cite{schuler2011non, kee2011modeling} for processing complex spatially-variant blur. 
Data-driven learning-based models~\cite{liang2021swinir,wang2022uformer}, which can be plugged directly into existing end-to-end frameworks of lens design~\cite{sun2021end,9919421,yang2023curriculum}, have demonstrated more powerful abilities in image recovery, but may hit a bottleneck when the training data is scarce. 

This motivates us to break the bottleneck by utilizing PSFs of the optical system in a learning-based model. 
The PSFs of $n$ sampled FoVs of the applied MIIP can be calculated based on the wavefront as depicted in Eq.~(\ref{eq:zernike}):
\begin{equation}
\label{eq:psf}
{K_i}(x,y) = \mathcal{S}({\Phi_i}(x',y')), i= 1,2,3, \cdots, n.
\end{equation}
The $\mathcal{S}(\cdot)$ denotes scalar diffraction integral (refer to~\cite{huggins2007introduction} for more details).
Previous methods tend to use the kernel size of $K_i$ to guide the network~\cite{jiang2023annular} or refine the ill-posed term in deconvolution through a learning-based model~\cite{lin2022non}. 
Although these attempts can improve the recovery benefiting from the applications of $K_i$, the spatial intensity distribution of PSFs is not fully exploited to guide the deep feature extraction of the general image recovery paradigm.

To this intent, we propose a PSF representation method to produce a PSF map, containing both intensity and size distributions of PSF kernels, which is aligned with the image feature map. We first map the spatial PSFs $K_i\in\mathbb{R}^{k_i{\times}{k_i}{\times}3}$ into the image feature shape ${H{\times}{W}{\times}{C'}}$, where $k_i$ is the kernel size of PSF under the $i_{th}$ FoV and $C'$ denotes the channels of mapped PSFs, serving as an additional modality aligned with the aberration-image.
As previously shown in~\cite{li2021involution}, the spatial-to-channel arrangement helps transform spatially-variant kernels into a feature map.
Similarly, to produce a PSF feature map, the spatial PSFs $K_i$ are arranged into the channel dimension.
Concretely, for a pixel at $(x,y)$ of the image, we first calculate the vector $\overrightarrow{p}$ from the image center $(0,0)$ to $(x,y)$, and define the vertical unit vector$\overrightarrow{a}(1,0)$.
The PSF of the corresponding FoV is located by $\left| \overrightarrow{p} \right|/\max \left( \left|{\overrightarrow{p}} \right| \right)$, and rotated by the angle $\arccos \left( \frac{\vec{p}\cdot \vec{a}}{\left| \vec{p} \right|\cdot \left| \vec{a} \right|} \right)$, producing the PSF $K_{x,y}$ of the pixel.  
For memory-friendly computation, we pad all $K_{x,y}$ into unified size of $\max\limits_{i}k_i$ and compress them into ${k'\times{k'}\times1}$ via adaptive average pooling:
\begin{equation}
\label{eq:pad}
\hat{K_{x,y}} = {\rm{AveragePool}}({\rm{padding}}{(K_{x,y})}),
\end{equation}
where the choice of compressed size $k'$ is ablated in Sec.~\ref{sec:ablation}. The $\hat{K_{x,y}}$ is then reshaped into ${1\times1\times(k'^2)}$ and inserted into each pixel to produce the PSF feature map $x_{int}\in\mathbb{R}^{H\times{W}\times(k'^2)}$.
In addition, considering the lost PSF size information during compressing, we also generate the size distribution map $x_s\in\mathbb{R}^{H\times{W}\times3}$ of RGB channels, where the value of each pixel represents the kernel size.
Finally, the PSF map $x_{psf}$ is produced via Eq.~(\ref{eq:compress}):
\begin{equation}
\label{eq:compress}
x_{psf} = {\rm{Concat}}{(x_{int}, x_{s})},
\end{equation}
and the visualized PSF map is shown in the appendix. 
PSF map is an aligned modality of the aberration image characterizing the image degradation over FoVs, based on which we design a PSF-aware transformer, as described in the next subsection. 

\subsection{PSF-aware Feature Modulator}
\label{sec:pfm}
CNN layers have shown impressive abilities of local feature extraction, but are restricted to the fixed spatially-invariant kernels.
However, the mathematical imaging model in Eq.~(\ref{eq:basic2}) reveals that the aberration-induced blur is only generated by convolution with spatially-variant PSF kernels, whose inverse solution cannot be modeled by the fixed convolution kernels~\cite{chen2021extreme_quality, chen2022computational_mass}. 

To extract adaptive image features with the guidance of spatially-variant PSF kernels, we propose the PSF-aware Feature Modulator (PFM), as shown in the lower left of Fig.~\ref{fig:part}. 
PFM builds on the idea of filter adaptive convolution~\cite{li2021involution, jiang2022fast}, where a kernel map of ${H{\times}{W}{\times}(Ck^2)}$ is predicted from feature map of ${H{\times}{W}{\times}{C}}$. 
Differently, in PFM, the kernel map $x_{kernel}$ is predicated on the features of PSF map $x_{psf}$, which has been compressed into a similar form as $x_{kernel}$.  
We first apply $E_{psf}$ as a $3\times{3}$ convolution layer to extract features of PSF map $x_{psf}$, as depicted in Eq.~(\ref{eq:E}):
\begin{equation}
\label{eq:E}
x'_{psf} = E_{psf}(x_{psf}),
\end{equation}
where $x'_{psf}\in\mathbb{R}^{H\times{W}\times{C}}$ is the extracted PSF feature map. 
Then, a lightweight kernel predictor composed of several convolution layers is proposed to output the kernel map $x_{kernel}\in\mathbb{R}^{H{\times}{W}{\times}(Ck^2)}$ based on $x'_{psf}$, as in Eq.~(\ref{eq:kernelmap}):
\begin{equation}
\label{eq:kernelmap}
x_{kernel} = P(x'_{psf}).
\end{equation}
To reduce the memory cost and inference latency of kernel prediction, the predictor $P$ computes the kernel map on the downsampled features (by $4{\times}4$ average pooling).
Benefiting from that the PSF map shares a similar form with the kernel map, we further simplify the $P$ where only one Max Pooling layer and one residual block of $1{\times}1$ convolution layers are applied, to predict the kernel map $x'_{kernel}\in\mathbb{R}^{\frac{H}{8}{\times}\frac{W}{8}{\times}(Ck^2)}$ in a smaller resolution.
The final kernel map $x_{kernel}$ is then obtained by ${\times}8$ upsampling via bilinear interpolation.
Finally, we reshape the $x_{kernel}$ into a list of per-pixel kernels of $k{\times}{k}{\times}{C}$ and apply them to the corresponding pixels of image feature $x'_{img}$. PFM attempts to model the inverse process of the aberration-induced blur, \ie~deconvolution, which promotes the dynamic feature extraction of the aberration-image.

\subsection{PSF-aware Mix-Attention Block}
\label{sec:pmab}
We put forward the PSF-aware Mix-Attention Block (PMAB) as the basic unit of our PSF-aware transformer, to process aberration images assisted with the PSF map, as shown at the middle bottom of Fig.~\ref{fig:part}. 
The Window-based Multi-head Self-Attention (W-MSA) of the Swin-T block~\cite{liang2021swinir} is first adopted to be the baseline attention mechanism for modeling spatially-variant convolution and long-range dependency, which is also important for stable training of the network. 

To address the drawback of fixed window size in vanilla W-MSA, we further propose the PSF-aware Varied-Size Attention (P-VSA), shown on the lower right of Fig.~\ref{fig:part}.
The vanilla varied-size attention~\cite{zhang2022vsa} in high-level tasks predicts the sizes and locations of the windows from input features for computing self-attention on dynamic windows. 
Meanwhile, the kernel sizes of PSFs in different FoV regions reveal the severity of aberration-induced blur, which is relevant to the calculation of window-based self-attention.
To better adaptively modulate the windows according to the PSF kernels, we make use of the PSF map features $x'_{psf}$ to generate PSF-aware varied-size windows.
Concretely, the scale $S$ and offset $O$ of the varied-size windows are predicated on $x'_{psf}$ by the Window Transform block, which is composed of a $1\times{1}$ convolution layer.
Then, we sample the projected key and value tokens $K,V$ of image features $x'_{img}$ based on the transformed window to obtain $K_{vs},V_{vs}$.
The cross-attention is computed between query $Q$ of the default window and $K_{vs},V_{vs}$.
The operation of P-VSA can be expressed as:
\begin{equation}
\label{eq:QKV}
Q,K,V = {\rm{Linear}}{({\rm{WinPar}}(x'_{img}))},
\end{equation}
\begin{equation}
\label{eq:win}
S,O = {\rm{WinTrans}}(x'_{psf}),
\end{equation}
\begin{equation}
\label{eq:sample}
K_{vs},V_{vs} = {\rm{Sample}}(K,S,O),{\rm{Sample}}(V,S,O)
\end{equation}
\begin{equation}
\label{eq:atten}
{\rm Attn}(Q, K_{vs}, V_{vs}) = {\rm Softmax}(\frac{QK_{vs}^{\top}}{\sqrt{d}})V_{vs},
\end{equation}
where $\rm{WinPar}$ denotes the window partition operation of Swin-T and $d$ is the dimension of tokens. 

Additionally, some works~\cite{li2023efficient, chen2205activating} apply channel-attention-based convolution blocks in parallel with the self-attention to enhance the representation ability of the network.
We insert the proposed PFM to PMAB in the same parallel way, where the filter adaptive convolution mechanism can better model the spatially-variant blur compared to channel-attention-based convolution. 

Finally, PMAB is the mixing of W-MSA and P-VSA with a parallel $1{\times}1$ PFM.
For the self-attention module, the image feature map $x'_{img}$ is equally split along the channel dimension and processed by parallel W-MSA and P-VSA, then concatenated along the channel dimension again.
The modulated feature map by parallel PFM is multiplied by a constant $\alpha$, to be added to the result of self-attention and the original feature map as common practice for stable training~\cite{chen2205activating}.
The whole process of PMAB is computed as:
\begin{equation}
\label{eq:split}
{x'}^{(1)}_{img}, {x'}^{(2)}_{img}={\rm{Split}}({x'}_{img}),
\end{equation}
\begin{equation}
\label{eq:pmab1}
x_{attn} = {\rm{Concat}}({\rm{M\text{-}WSA}}({x'}^{(1)}_{img}),{\rm{P\text{-}VSA}}({x'}^{(2)}_{img}, x'_{psf})),
\end{equation}
\begin{equation}
\label{eq:pmab2}
x_{mix} = x_{attn} + \alpha{\rm{PFM}}({x'}_{img}, {x'}_{psf}) + {x'}_{img},
\end{equation}
\begin{equation}
\label{eq:pmab3}
y = x_{mix} + {\rm{FFN}}(x_{mix}),
\end{equation}
where ${\rm{FFN}}$ is a common Feed Forward Network composed of a LayerNorm and a Multi-Layer Perceptron (MLP) layer.

\subsection{PSF-aware Aberration-image Recovery Transformer}
\label{sec:network}
Most previous networks~\cite{peng2019learned, chen2021extreme_quality} for aberration corrections often utilize the architecture of image deblurring methods, \ie~U-Net.  
However, for MPIP images with a high resolution (\textit{e.g.}, $3K$), the U-Net methods incur unacceptable computational costs due to the large image sizes at shallow layers.
Differently, we look into the tasks from the perspective of image super-resolution, which processes image features with low resolution and reconstructs the high-quality image via an upsampling module. 
The aberration-induced blur brings aliasing between pixels and losses of image details, which can also be interpreted as ``low resolution''. 
Thereby, as shown in Fig.~\ref{fig:part}, the PSF-aware Aberration-image Recovery Transformer (PART) is set up based on the structure of SwinIR~\cite{liang2021swinir} and our proposed PSF-aware mechanisms.

A Task-Processing module is first applied to transform the input image and PSF map to a small spatial size, where pixel-unshuffle~\cite{wang2021real} is leveraged for AC and no operation is entailed for SR$\&$AC. 
The PSF map is also concatenated with the aberration image as the input of the network. 
More precisely, PART contains three parts.
(1)~A feature extraction layer converts the input to image feature maps via a $3{\times}3$ convolution.
(2)~The representation learning stage applies stacks of transformer-based blocks ending with a convolution layer to enrich the learned degradation information of aberration-induced blur progressively.
We design the PSF-aware Residual Transformer Block (PRTB) with several PMAB layers and a convolution layer.
The PFM is inserted into each PRTB to modulate the learned features and model the inverse process of the aberration-induced blur.
We also implement PFM at the beginning and end of the representation learning stage, for adaptive feature extraction and feature fusion based on PSF information.
(3)~The image reconstruction module further fuses the extracted deep features and recovers a high-quality image with higher resolution. 
With PART, we can recover a high-quality aberration-free image $x_{hq}$ from either a high-resolution aberration image $x_{ab}$ or a low-resolution one $x_{lq}$, providing a general solution to AC and SR$\&$AC:
\begin{equation}
\label{eq:part}
{x^{AC}_{hq}} = {\rm{PART}}(x_{ab}), 
{x^{SR\&AC}_{hq}} = {\rm{PART}}(x_{lq}).
\end{equation}

\section{Experiments and Results}
\label{sec:exp}
\input{Tex_content/experiments_revised}

\section{Conclusion and Discussion}
\input{Tex_content/conclusion_revised}

{\small
\bibliographystyle{IEEEtran}
\bibliography{bib}
}

\input{Tex_content/appendix_revised}

\end{document}

%% file: Tex_content/abstract_revised.tex
High-quality panoramic images with a Field of View (FoV) of 360{\textdegree} are essential for contemporary panoramic computer vision tasks. However, conventional imaging systems come with sophisticated lens designs and heavy optical components. This disqualifies their usage in many mobile and wearable applications where thin and portable, minimalist imaging systems are desired. In this paper, we propose a Panoramic Computational Imaging Engine (PCIE) to achieve minimalist and high-quality panoramic imaging. With less than three spherical lenses, a Minimalist Panoramic Imaging Prototype (MPIP) is constructed based on the design of the Panoramic Annular Lens (PAL), but with low-quality imaging results due to aberrations and small image plane size. We propose two pipelines, \ie~Aberration Correction (AC) and Super-Resolution and Aberration Correction (SR$\&$AC), to solve the image quality problems of MPIP, with imaging sensors of small and large pixel size, respectively. To leverage the prior information of the optical system, we propose a Point Spread Function (PSF) representation method to produce a PSF map as an additional modality. A PSF-aware Aberration-image Recovery Transformer (PART) is designed as a universal network for the two pipelines, in which the self-attention calculation and feature extraction are guided by the PSF map. We train PART on synthetic image pairs from simulation and put forward the PALHQ dataset to fill the gap of real-world high-quality PAL images for low-level vision. A comprehensive variety of experiments on synthetic and real-world benchmarks demonstrates the impressive imaging results of PCIE and the effectiveness of the PSF representation. We further deliver heuristic experimental findings for minimalist and high-quality panoramic imaging, in terms of the choices of prototype and pipeline, network architecture, training strategies, and dataset construction. Our dataset and code will be available at \url{https://github.com/zju-jiangqi/PCIE-PART}.

%% file: Tex_content/Introduction_revised.tex
\begin{figure}[!t]
  \centering
  \includegraphics[width=0.85\linewidth]{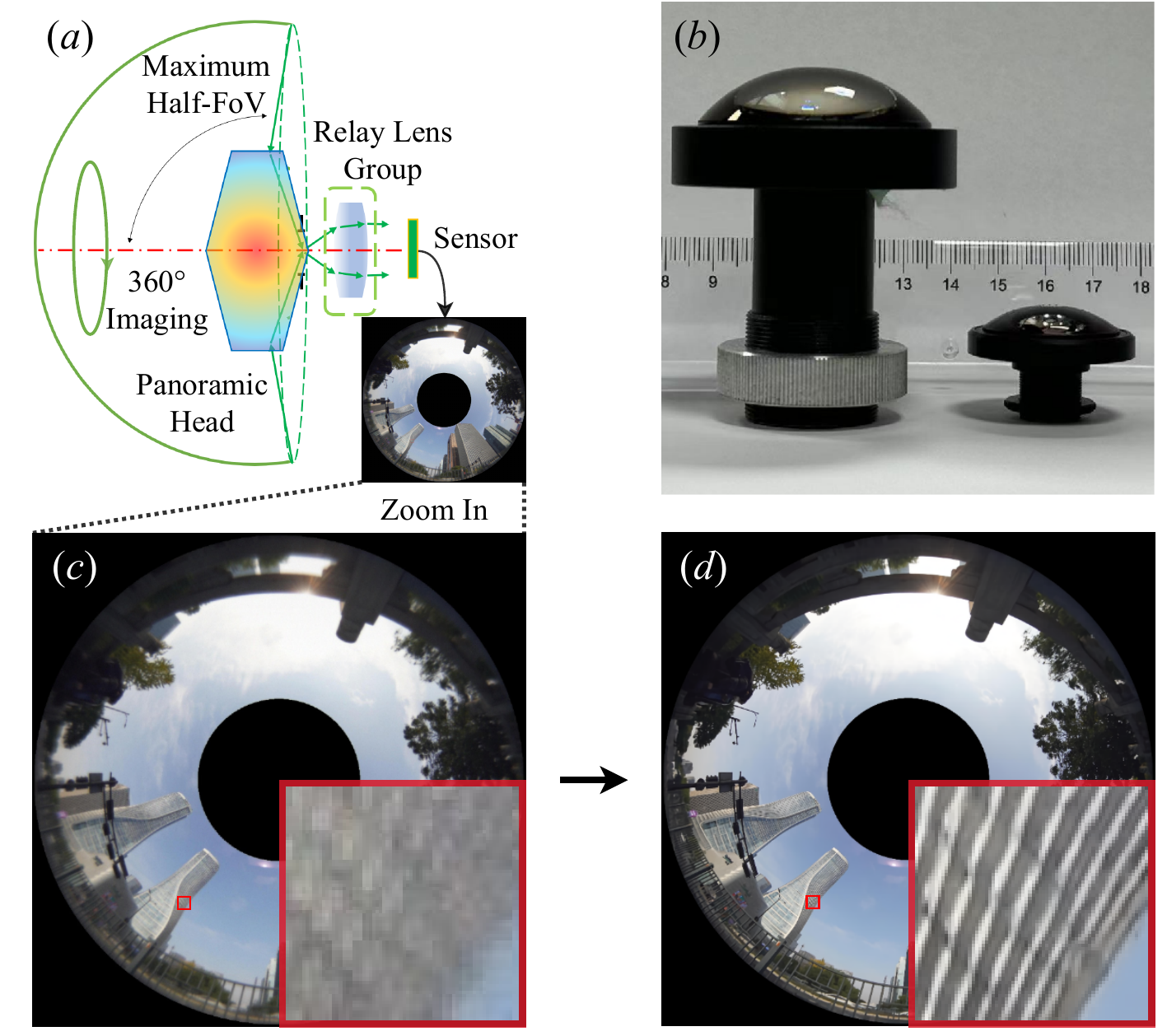}
  \caption{Illustration of the proposed MPIP and its key issue of low image quality, which is properly addressed with PSF-aware transformer: PART. (a)~Minimalist Panoramic Imaging Prototype (MPIP); (b)~Comparison between real products of conventional panoramic imaging systems and PAL-based MPIP; (c)~Low-quality image captured by MPIP. (d)~High-quality image recovered by PART. In this way, we realize minimalist and high-quality panoramic imaging with PSF-aware transformers.}
  \label{fig:intro}
\end{figure}

\IEEEPARstart{I}{mage} processing of panoramic images with an ultra-wide Field of View (FoV) of 360{\textdegree} is growing popular for achieving a holistic understanding of the entire surrounding scene~\cite{yang2021context,tateno2018distortion,shen2022panoformer,liao2022cylin}.
While the 360{\textdegree} panoramas suffer from inherent defects of low angular resolutions and severe geometric image distortions, a variety of low-level vision works is conducted in terms of image super-resolution~\cite{yoon2022spheresr,yu2023osrt,sun2023opdn} and image rectification~\cite{yang2022fishformer,yang2021progressively,yang2023dual}, to produce high-quality panoramas for photography and down-stream tasks.
However, the image blur caused by optical aberrations of the applied lens is seldom explored.

Most contemporary works on panoramic images are based on the common sense that the optical system is aberration-free where the imaging result is clear and sharp.
While widely applied, conventional panoramic optical systems, come with notoriously sophisticated lens designs, composed of multiple sets of lenses with complex surface types~\cite{gao2022review, Cheng2016DesignOA, gao2021design}, to reach high imaging quality.
However, this is not often the case as the demand for thin, portable imaging systems, \ie, Minimalist Optical Systems (MOS), grows stronger in mobile and wearable applications~\cite{jiang2022computational}. 
Without sufficient lens groups for aberration correction, the aberration-induced image blur is inevitable for MOS. 
In this case, the imaging quality drops significantly and often catastrophically, and the unsatisfactory imaging performance disqualifies its potential usage in upper-level applications.
This leads to an appealing issue and we ask if we may strike a fine balance between high-quality panoramic imaging and minimalist panoramic optical systems.

With the rapid development of digital image processing, Computational Imaging (CI) methods for MOS~\cite{peng2016diffractive,peng2019learned,li2021universal} appear as a preferred solution to this issue. 
These methods often propose optical designs with few necessary optical components to meet the basic demands of specific applications, \eg, the FoV, depth of field, and focal length, followed by an image post-processing model to recover the aberration-image.
Recent research works~\cite{sun2021end,9919421,yang2023curriculum} further design end-to-end deep learning frameworks for joint optimization of optical systems and image recovery networks.
In this paper, based on the idea of computational imaging, we propose \textit{Panoramic Computational Imaging Engine (PCIE)}, a framework for minimalist and high-quality panoramic imaging, to solve the trade-off between high-quality panoramic imaging and minimalist panoramic optical systems as a whole, without sitting on only one of its sides.

Motivated by modern panoramic lens designs~\cite{Cheng2016DesignOA, Zhang2020DesignOA,wang2022design}, PCIE builds on a Minimalist Panoramic Imaging Prototype (MPIP) shown in Fig.~\ref{fig:intro}(a), which is composed of an essential panoramic head for 360{\textdegree} panoramic imaging and a relay lens group for aberration correction.
In specific, we select the structure of Panoramic Annular Lens (PAL)~\cite{greguss1986panoramic,powell1994panoramic}: a more compact solution to 360{\textdegree} panoramas, as an example for MPIP, where a catadioptric PAL head is equipped to replace the complex lens group~\cite{gao2022review} in the conventional fisheye lens~\cite{Thibault2005EnhancedOD, Geng2017OpticalSD}.
To achieve a minimalist design, the proposed MPIP is composed of $1$ spherical lens for the PAL head, and $1$ or $2$ simple spherical lenses for the relay lens group, which can image over 360{\textdegree} FoV with only $40\%$ of the numbers of lenses and $60\%$ of the volumes of conventional panoramic imaging systems, as shown in Fig.~\ref{fig:intro}(b). 
However, as illustrated in Fig.\ref{fig:intro}(c), the uncorrected optical aberrations and the limited image plane size lead to the image corruptions, \ie, aberration-induced spatially-variant image blur and low imaging resolution.

To address the issues of MPIP, engaged with the information of Point Spread Function (PSF) from optical design, we propose \textit{PSF-aware Aberration-image Recovery Transformer (PART)}: a transformer-based low-quality image recovery paradigm for MPIP. 
Different from previous transformer baselines, \eg, SwinIR~\cite{liang2021swinir}, PART exploits the PSF, the forward function characterizing the aberration-induced blur, to attain better results. 
A PSF representation method is delivered to represent PSF kernels in the form of a feature map, which serve as an additional modality for the network.
Based on the representation, we design two PSF-aware mechanisms inspired by the physical meanings of the aberration-induced blur. 

Specifically, the PSF-aware Feature Modulator (PFM) builds on the idea of modeling the inverse process of degradation convolution of PSFs, where pixel-adaptive convolution kernels are learned from the PSF representation to modulate the feature map gradually during recovery.
PFM is a plug-and-play PSF-aware mechanism that can be inserted into other recovery models.
In addition, PSF-aware Mix-Attention Block (PMAB) is proposed as the basic unit of PART, which comprises:
(1)~Vanilla window attention of SwinIR~\cite{liang2021swinir} for capturing long-range dependency;
(2)~PSF-aware Varied-Size Attention (P-VSA), where diverse windows of varied sizes and locations are learned from the PSF representation to provide dynamic receptive fields, motivated by the varied PSF sizes in different FoVs;
(3)~PFM of small kernel size for enhancing the feature extraction of local details.
With PART, the low-quality image captured by MPIP can be smoothly recovered (see Fig.~\ref{fig:intro}(d)) for minimalist and high-quality panoramic imaging.

To facilitate the training of PART, wave-based imaging simulation with random perturbation~\cite{jiang2023annular} is utilized for generating clear-blur image pairs. 
To fill the gap of ground-truth images of PAL, we record a high-quality PAL images dataset named \textit{PALHQ} through a well-designed PAL in varied scenes. 
Based on PALHQ, we set up two tasks to formalize the key issue of low-quality MPIP images:
(1)~\textit{Aberration Correction (AC)} of high-resolution images taken by sensors with small pixel size,
and (2)~\textit{Super-Resolution and Aberration Correction (SR$\&$AC)} of low-resolution images from sensors with large pixel size.
Then, representative models of image super-resolution (SR)~\cite{liang2021swinir,wang2018esrgan,li2023efficient,li2021efficient, chen2205activating, lim2017enhanced, zhang2018image}, image deblurring (Deblur)~\cite{zamir2022restormer,wang2022uformer,chen2021hinet,chen2022simple}, and image restoration with PSF-aware mechanisms (PSF-aware)~\cite{jiang2023annular} are evaluated, where PCIE enables all models to produce impressive panoramic imaging results.

Furthermore, we manufacture an MPIP sample with better image quality and capture the real-world dataset \textit{RealMPAL} to benchmark models on real-world scenes. 
Experimental results reveal that PFM enhances the performance of the baselines (see Fig.~\ref{fig:pfm}) and PART sets the state of the art on both synthetic and real-world benchmarks, where the PSF representation plays a significant role to enable effective PSF-aware mechanisms. 
We also conduct extensive experiments to investigate the potential of GAN-based training strategies and the effectiveness of PALHQ in PCIE. 
The generative model appears to be competitive for generating more realistic details if the artifacts can be well suppressed.
Additionally, PALHQ serves as the cornerstone of PCIE for training a robust model for annular images.
At a glance, we deliver the following contributions:

\begin{figure}[!t]
  \centering
  \includegraphics[width=1\linewidth]{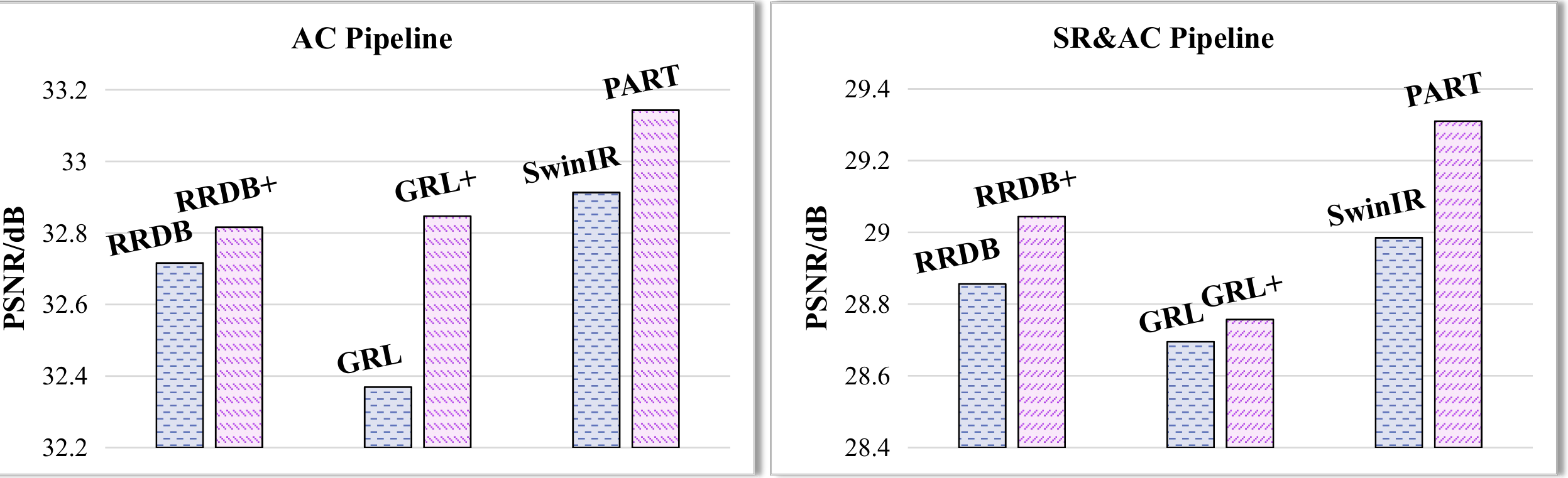}
  \caption{The proposed plug-and-play PSF-aware mechanism, PFM, consistently and significantly improves the performance of several baseline models in two pipelines. ``+'' means the model inserted the PFM in the same way as PART.}
  \label{fig:pfm}
\end{figure}

%% file: Tex_content/contribution_revised.tex
\begin{compactitem}
    \item We propose the Panoramic Computational Imaging Engine (PCIE), a novel framework for minimalist and high-quality panoramic imaging, as shown in Fig.~\ref{fig:overview}, where a Minimalist Panoramic Imaging Prototype (MPIP) is designed for 360{\textdegree} panoramic imaging with an essential panoramic head and simple relay lens group. 
    \item We raise two pipelines to process low-quality MPIP images: Aberration Correction (AC) and Super-Resolution and Aberration Correction (SR$\&$AC). The real-world panoramic image datasets PALHQ and RealMPAL of high-quality and low-quality are recorded respectively for benchmarking the two pipelines, which are the first real-world PAL datasets for low-level vision.
    \item We design a PSF representation method to represent the intensity and size distributions of PSF kernels in the form of a feature map, \ie, PSF map, which serves as an additional modality for the pipelines.
    \item We further introduce the PSF-aware Aberration-image Recovery Transformer (PART) to process the low-quality images of MPIP, where the PSF-aware mechanisms guided by the PSF map are explored to enhance the recovery performance.
\end{compactitem}

The experimental exploration of PCIE provides heuristic findings in terms of optical design, network architecture, training strategies, and dataset construction. We hope that PCIE can bring inspiration in both hardware system and algorithm aspects, for minimalist and high-quality panoramic imaging.

%% file: Tex_content/related_work_revised.tex
\subsection{Image Processing of Panoramic Images}
Recent research interest in panoramic images is booming for immersive visual experiences~\cite{gao2022review,ai2022deep_omnidirectional}. 
Semantic segmentation~\cite{yang2021context,zhang2022bending}, depth estimation~\cite{tateno2018distortion,shen2022panoformer}, and visual Simultaneous Localization and Mapping (SLAM)~\cite{chen2021panoramic_SLAM,wang2022lf} are widely explored on panoramic images for a holistic understanding of the surrounding scene. 
To this intent, high-quality panoramic images are urgently required for robust performance. A considerate amount of work is conducted to improve the image quality of panoramic images, such as super-resolution~\cite{yoon2022spheresr,yu2023osrt,sun2023opdn} and distortion correction~\cite{yang2022fishformer,yang2021progressively,yang2023dual}.

However, the above image processing of panoramic images is based on the aberration-free images captured by the conventional panoramic lens, where multiple sets of lenses with complex surface
types~\cite{gao2022review, Cheng2016DesignOA, gao2021design} are applied for high-quality imaging. 
This work focuses on capturing panoramic images with a Minimalist Optical System (MOS) composed of much fewer lenses for volume-limited applications, where we process the aberration images via computational methods.   

\subsection{Computational Imaging for Minimalist Optical System}
The aberration-induced image blur is inevitable for MOS due to insufficient lens groups for aberration correction.
Recently, computational imaging methods for MOS appear as a preferred solution to this issue, where optical designs with few necessary optical components are equipped with image recovery pipelines for both minimalist and aberration-free imaging~\cite{peng2019learned,li2021universal,schuler2011non}. 
Some research works~\cite{sun2021end,liu2021end,cote2022differentiable} even design end-to-end deep learning frameworks for joint optimization of MOS and post-processing networks to achieve the best match between MOS and recovery models to further improve the final imaging performance.

However, computational imaging for minimalist panoramic systems is scarcely explored.  
In a preliminary study, Jiang~\textit{et al.}~\cite{jiang2023annular} propose an Annular Computational Imaging (ACI) framework to break the optical limit of minimalist Panoramic Annular Lens (PAL), where the image processing is conducted on unfolded PAL images.
To further develop a general framework for minimalist and high-quality panoramic imaging, this work fills the gap in high-quality PAL image datasets and designs the Panoramic Computational Imaging Engine (PCIE) directly on annular PAL images, in terms of both optical design and aberration-images recovery.

\subsection{Image Recovery of Aberration Images}
The aberration-induced blur is always spatially-variant, \ie~ Linear Shift Variant (LSV), due to the uneven thicknesses of the lenses.
Several efforts have been made for the LSV system, spanning path-wise restoration~\cite{trussell1978image, kim2002curve}, experimental PSFs calibration and non-blind deconvolution~\cite{schuler2011non, kee2011modeling}, and low-rank decomposition~\cite{xue2022deep, denis2015fast}, based on the degradation model of aberration-images~\cite{wiener1949extrapolation, richardson1972bayesian,lucy1974iterative}. 

Recent works tend to adopt the data-driven learning-based image restoration networks~\cite{sun2015learning,chen2022simple}.
These methods typically use a U-shaped network with an encoder-decoder structure~\cite{peng2019learned,chen2021extreme_quality, chen2022computational_mass} to achieve more efficient and robust recovery results, which can also be easily inserted into an end-to-end framework for joint optimization.
To break the bottleneck of data-driven methods under scarce data, the PSF information is explored to design physical-informed networks, where model-based methods are characterized by Convolutional Neural Networks (CNNs) for learning ill-posed terms~\cite{jiang2022computational,lin2022non}.
Explorations have also been made in~\cite{liang2021swinir, zamir2022restormer, wang2022uformer, ma2023learning} to apply transformers for solving the inverse problem, leveraging its strong long-range modeling capabilities.

Differently, we make a pioneering effort and investigate the potential of transformer-based SR models in aberration correction rather than conventional Deblur models. 
Then, the PSFs are transformed into an additional modality of the aberration image, based on which we design PSF-aware mechanisms for achieving better results.
The proposed PSF-aware Aberration-image Recovery Transformer (PART) is a successful attempt to engage PSF information in the representation learning stage of SR models for recovering aberration images.

The overview of PCIE is shown in Fig.~\ref{fig:overview}. It provides a powerful framework for minimalist and high-quality panoramic imaging, where optical design (detailed in Sec.~\ref{sec:MPIP}) and learning-based model (presented in Sec.~\ref{sec:part}) are intertwined to achieve impressive imaging results.

%% file: Tex_content/experiments_revised.tex
We conduct a comprehensive set of experiments to evaluate the proposed PCIE for minimalist and high-quality panoramic imaging.
We first describe the implementation details of our work in Sec.~\ref{sec:detail}.
The PCIE under different recovery models is then evaluated on both synthetic (Sec.~\ref{sec:syn}) and real (Sec.~\ref{sec:real}) datasets. 
We further investigate the GAN-based training strategies for PCIE in Sec.~\ref{sec:gan}.
At last, in Sec.~\ref{sec:data} and Sec.~\ref{sec:ablation}, ablation studies on training datasets and the architecture of PART are conducted.

\subsection{Implementation Details}
\label{sec:detail}

\begin{table*}[!t]

	\renewcommand\arraystretch{1.2}
	\setlength{\fboxrule}{0pt}
		\captionsetup{font={footnotesize}}

	\begin{center}
		\input{Table/synall}
	\end{center}

\end{table*}

\noindent\textbf{Synthetic Datasets.}
We apply the collected PALHQ dataset for training and evaluation.
Based on PALHQ, the aberration images of two prototypes, \ie~PALHQ-SynMPIP-P1 and PALHQ-SynMPIP-P2, are generated by the simulation model of Eq.~(\ref{eq:basic2}).
Following~\cite{jiang2023annular}, we set the random range of disturbance as $25\%$ and generate $10$ virtual MPIP samples for the training set ($500$ images) and $4$ for the validation set ($50$ images) to simulate the synthetic-to-real gap.
For image sensors, the MV-SUA1600C camera with a pixel size of $1.34{\mu}m$ and the MV-SUA133GC camera with a pixel size of $4{\mu}m$ are applied for the AC and SR$\&$AC pipelines, respectively, where the ISP and wave response of them are simulated in the data generation. 
In addition, we use $\times3$ bicubic downsampling to produce low-resolution aberration images for SR$\&$AC, considering the sensors' pixel sizes.

\noindent\textbf{Real-world Datasets.}
As shown in Fig.~\ref{fig:pal}, with only one more simple lens, the MPIP-P1 reveals much better image quality, which relieves the burden of the post-image processing pipelines. 
We manufacture MPIP-P1 and use it to record the RealMPIP3K-AC ($58$ images with a resolution of $2912\times2912$) and RealMPIP1K-SR$\&$AC ($64$ images with a resolution of $992\times992$) with two cameras respectively, to provide real-world MPIP aberration-images for evaluating two pipelines of PCIE.
We test models trained on PALHQ-SynMPIP-P1 (AC) and PALHQ-SynMPIP-P1 (SR$\&$AC) with RealMPIP3K-AC and RealMPIP1K-SR$\&$AC respectively. 

\noindent\textbf{Evaluation Metrics.}
For synthetic datasets with ground truth, PSNR and SSIM~\cite{wang2004image} are employed to evaluate the fidelity of the recovery results, whereas LPIPS~\cite{zhang2018unreasonable} and FID~\cite{heusel2017gans} are employed to evaluate the perceptual quality. 

For real datasets without reference clear image, we employ no-reference metrics, \ie~NIQE and BRISQUE, to evaluate the image quality of MPIP images in terms of natural images.
The qualitative visual results are also provided for an intuitive evaluation. 
However, the NIQE~\cite{mittal2012making} and BRISQUE~\cite{mittal2011referenceless} are built on the statistics of perspective natural images, which are challenging for assessing the MPIP images with the annular distribution of image content.
Considering the specific tasks of correcting optical aberrations, we define the Optical-based Image Quality Evaluator (OIQE) for credible evaluation, based on the Modulation Transfer Function (MTF) of the imaging system calculated by a set of testing checkerboard images. 

To be specific, we follow Spatial Frequency Response (SFR)~\cite{chen2022computational_mass} testing to calculate MTFs on image patches of ``knife-edge'' of different FoVs from different testing images. 
$MTF50$ and $MTFarea$ are used to characterize the MTF curves, where the former is the frequency when the MTF drops $50\%$ and the latter is the area under the MTF curve. 
We further define $OIQE50$ and $OIQEarea$ as the ratio of the average $MTF50$ and $MTFarea$ of the testing imaging pipeline to those of a well-designed panoramic imaging system. 
Accordingly, $OIQE$ is defined as:
\begin{equation}
\label{eq:oiqe}
OIQE = \frac{OIQE50+OIQEarea}{2},
\end{equation}
which measures the gap between the results of PICE and conventional panoramic lenses in terms of MTF.
OIQE is only applied in the AC pipeline due to its specific design for evaluating the ability of the model to remove aberration-induced blur. 

In addition, with the testing checkerboard images of OIQE, we generate the ground-truth images through edge extraction and re-coloring following~\cite{chen2021extreme_quality}, so that the PSNR and SSIM can be applied as metrics in this setting.

Finally, we conduct a user study as a subjective evaluation method. The results of the User Study (U.S.) will be presented as the percentage of times that each method's results were chosen as the best. 

The implementation details of the ground-truth generation pipeline and user study are depicted in the Appendix. Based on the above evaluation pipelines and metrics, a comprehensive evaluation of competitive recovery models on real-world datasets will be presented in Section~\ref{sec:real}.

\noindent\textbf{Compared Methods.}
For the AC pipeline, as shown in Table~\ref{tab:synac}, we compare PART with representative state-of-the-art SR models (RRDB~\cite{wang2018esrgan}, RCAN~\cite{zhang2018image}, EDSR~\cite{lim2017enhanced}, SwinIR~\cite{liang2021swinir}, EDT~\cite{li2021efficient}, HAT~\cite{chen2205activating}, and GRL~\cite{li2023efficient}),
along with Deblur methods (HINet~\cite{chen2021hinet}, NAFNet~\cite{chen2022simple}, Restormer~\cite{zamir2022restormer}, and UFormer~\cite{wang2022uformer}).
Image restoration models with PSF-aware mechanisms, \ie~RRDB+, GRL+, and PI$^2$RNet~\cite{jiang2023annular}, are also included in the comparison. 
Here, ``+'' means that the methods are inserted with the designed PFM, where we select RRDB and GRL as the classical CNN- and state-of-the-art transformer-based SR model to investigate the adaptability of PSF-aware mechanisms to different types of models.
For the PSF-aware methods in SR$\&$AC pipeline, only RRDB+ and GRL+ are selected due to the specific task requirement for super-resolution, as shown in Table~\ref{tab:synsrac}.

All the models are retrained on PALHQ-SynMPIP-P1 and PALHQ-SynMPIP-P2 with their original optimizers, learning rates, and schedulers, where the number of training iterations and the batch size are set the same as PART for a fair comparison.
Additionally, we apply task-processing for all the SR models the same as PART. 

\noindent\textbf{Training Details.}
The compressed kernel size $k'$ of the PSF map is set to $5$ in our experiments, where an ablation study is conducted in Sec.~\ref{sec:ablation}. In addition, we set the kernel size $k$ of PFM to $3$ considering the computational efficiency.
Following SwinIR~\cite{liang2021swinir}, the PRTB number, PMAB number, channel number, attention head number, and window size are generally set to $6$, $6$, $180$, $6$, and $8$, respectively.

PART is trained on L1Loss, while other loss functions are explored in Sec.~\ref{sec:gan}.
We train the models with the Adam optimizer with an initial learning rate of $2e{-}4$ and a batch size of $8$ on a single A800 GPU. For data augmentation, random crop, flip, and rotation are applied, where the ground-truth crop size is $256{\times}256$ for AC and $196{\times}196$ for SR$\&$AC to keep an image size of $64$ in the representation learning stage.
The number of training iterations is set to $200k$ and the learning rate is halved at $100k$, $160k$, $180k$, and $190k$. 

\begin{figure*}[!t]
  \centering
  \includegraphics[width=0.9\linewidth]{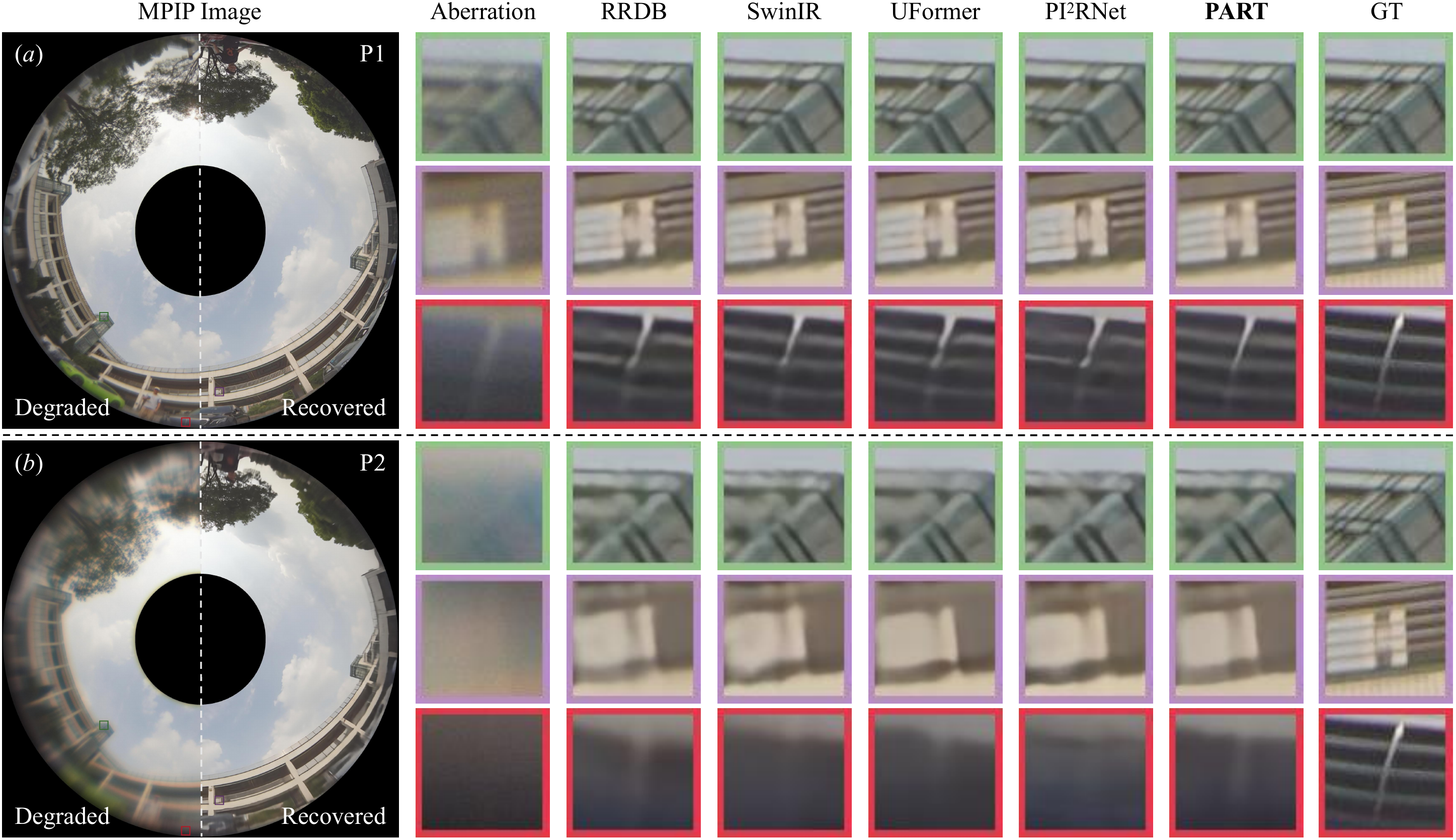}

  \caption{Qualitative results of representative models on synthetic benchmarks for AC. We zoom in on image patches of different FoVs to show the details. (a) Results on PALHQ-SynMPIP-P1, where aberration-free images are produced by all models and PART provides more visually pleasant and clearer details. (b) Results on PALHQ-SynMPIP-P2. PCIE delivers clear aberration-free images, yet, with heavily corrupted detailed textures due to severe optical aberrations.
  Enabling much higher imaging quality with only one more spherical lens, MPIP-P1 is a superior choice for PCIE.}
  \label{fig:q1}

\end{figure*}

\subsection{Experiments on Synthetic Datasets}
\label{sec:syn}

\noindent\textbf{AC Pipeline.}
Table~\ref{tab:synac} shows numerical results of PCIE under different image recovery models on synthetic benchmarks of AC. 
Considering that the performance of NAFNet and Restormer is sensitive to input resolution~\cite{chu2022improving,beyer2023flexivit}, the cropping testing strategy is applied for the two models (\ie, NAFNet* and Restormer*), which is depicted in the Appendix. 
We also present visual results of representative methods in Fig.~\ref{fig:q1}. 
PCIE with most models achieves PSNR over $32dB$ on PALHQ-SynMPIP-P1 and over $26dB$ on PALHQ-SynMPIP-P2, producing impressive panoramic imaging results via a minimalist optical system.
Compared to Deblur methods, SR methods overall deliver better results, illustrating the effectiveness of the SR framework in aberration correction. 
PSF-aware methods further outperform their baselines.
Precisely, PI$^2$RNet exceeds HINet, PART surpasses SwinIR, and RRDB+ and GRL+ outstrip their corresponding baselines by clear margins.
We find that the models based on the window-attention mechanism (SwinIR, EDT, HAT, UFormer, GRL, and our proposed PART) realize more competitive results than CNN-based models, where the window-based self-attention can better model spatially-variant blur. 
Yet, the state-of-the-art SR model GRL performs poorly on the benchmarks, which is attributed to the stripe-based attention being difficult to adapt to MPIP images with annular distributions.  

Overall, PART brings better results for PCIE, yielding state-of-the-art performance on two benchmarks, in terms of both fidelity-based metrics (PSNR and SSIM) and perceptual-based metrics (LPIPS and FID).
As all the methods produce aberration-free visual results with some lost textures and artifacts, the recovered image of PART shows more visually pleasant details, as shown in Fig.~\ref{fig:q1}(a), in all FoVs. 

Further, applied with only one more spherical lens, the PCIE results of MPIP-P1 outperform those of MPIP-P2 by a large margin. 
For example, the PSNR drops by $3.050dB{\sim}6.101dB$ when MPIP-P2 is equipped.
As shown in Fig.~\ref{fig:q1}(b), PCIE with MPIP-P2 delivers moderate clear aberration-free images.
Yet, suffering from severe aberrations, its detailed textures are heavily corrupted, especially for large FoVs.
In this sense, MPIP-P1 is a superior choice for PCIE to achieve minimalist and high-quality panoramic imaging.

\noindent\textbf{SR$\&$AC Pipeline.}
The quantitative evaluation of PCIE with the SR$\&$AC pipeline is shown in Table~\ref{tab:synsrac}.
Consistent with the observations in AC, the methods with window-based attention and PSF-aware mechanisms lead to better performance.
PART sets the state of the art in the SR$\&$AC task, achieving improvements compared against the second best, \eg~$0.266dB$ in PSNR, $0.0038$ in SSIM, LPIPS from $0.0714$ to $0.0681$ (about $5\%$), and FID from $9.709$ to $9.648$ (about $6\%$). 

\begin{figure}[!t]
  \centering
  \includegraphics[width=0.95\linewidth]{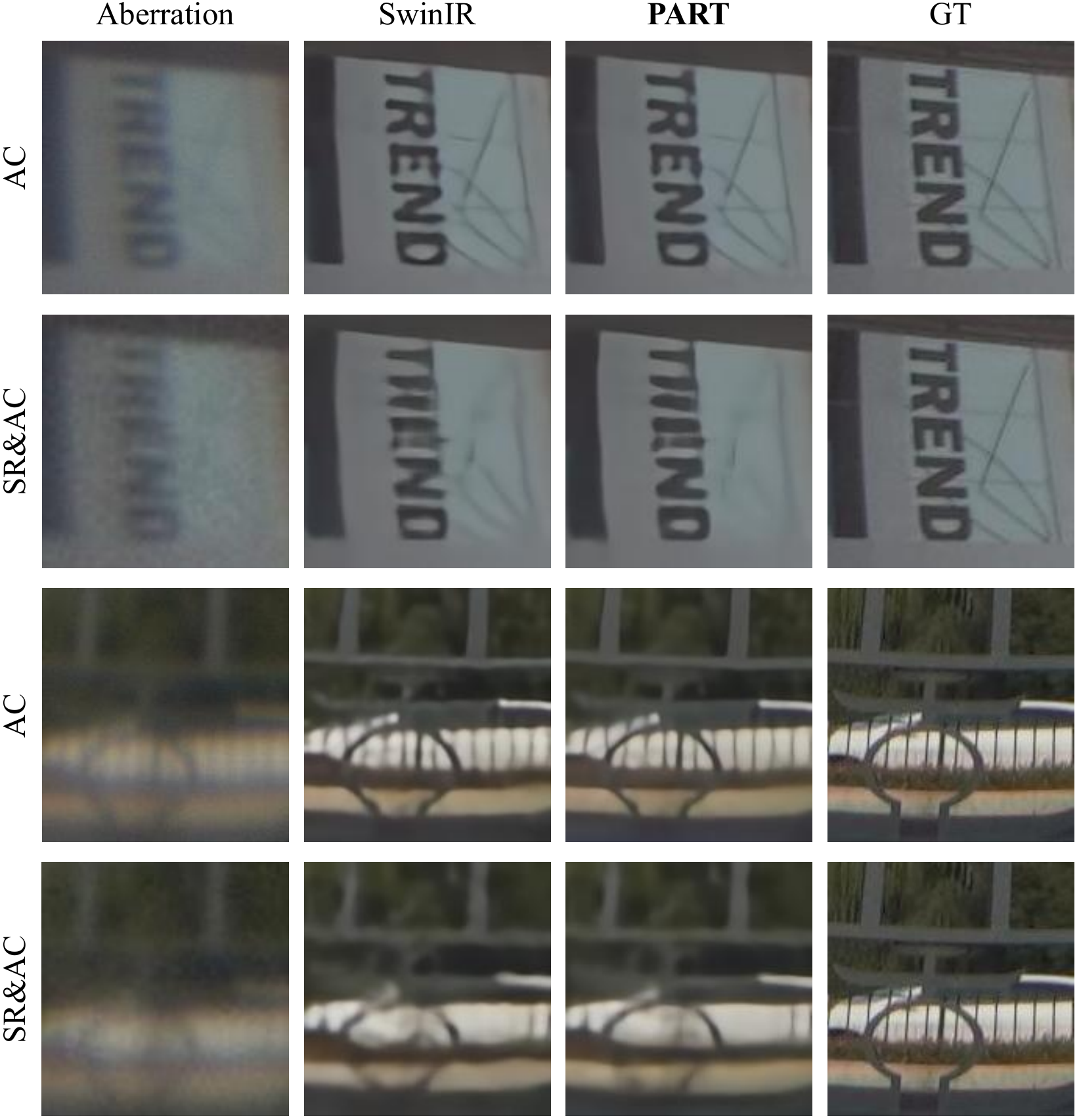}

  \caption{Comparison between AC and SR$\&$AC. The image patches are cropped from PALHQ-0532 (top) and PALHQ-0505 (bottom). We show the results of the proposed PART and its baseline SwinIR to illustrate the strengths of the AC pipeline, which produces richer and more realistic image details. }
  \label{fig:srac}

\end{figure}

Comparing SR$\&$AC (Table~\ref{tab:synsrac}) with AC (Table~\ref{tab:synac}), we observe that the loss of spatial resolution in aberration-images causes significant deterioration to the imaging quality of PCIE, \eg, to an amount of ${-}4.258dB{\sim}{-}3.674dB$ in PSNR. 
The visual quality comparison between the two pipelines is provided in Fig.~\ref{fig:srac}, where the imaging results of AC reveal richer and more realistic details.  
In this case, AC is a more competitive pipeline for reconstructing high-resolution aberration-free images, where the real sampled pixels of the sensor offer more convincing imaging features than super-resolved ones despite more aberration-induced blur.

\begin{table*}[!t]
    \begin{center}
        \caption{Quantitative evaluation of PCIE on real-world benchmarks RealMPIP. The OIQE and PSNR/SSIM of original aberration images are $55.22\%$ and $16.215dB/0.7995$, respectively. The PSNR and SSIM are calculated on the generated checkerboard image pairs. U.S. denotes the result of the user study. We also list the ranks on each metric in ``()'' and the Average Rank (A.R.) of each method for an intuitive evaluation.}

        \label{tab:real}
        \input{Table/table_real_revised}
    \end{center}

\end{table*}

\begin{figure}[!t]
  \centering
  \includegraphics[width=0.95\linewidth]{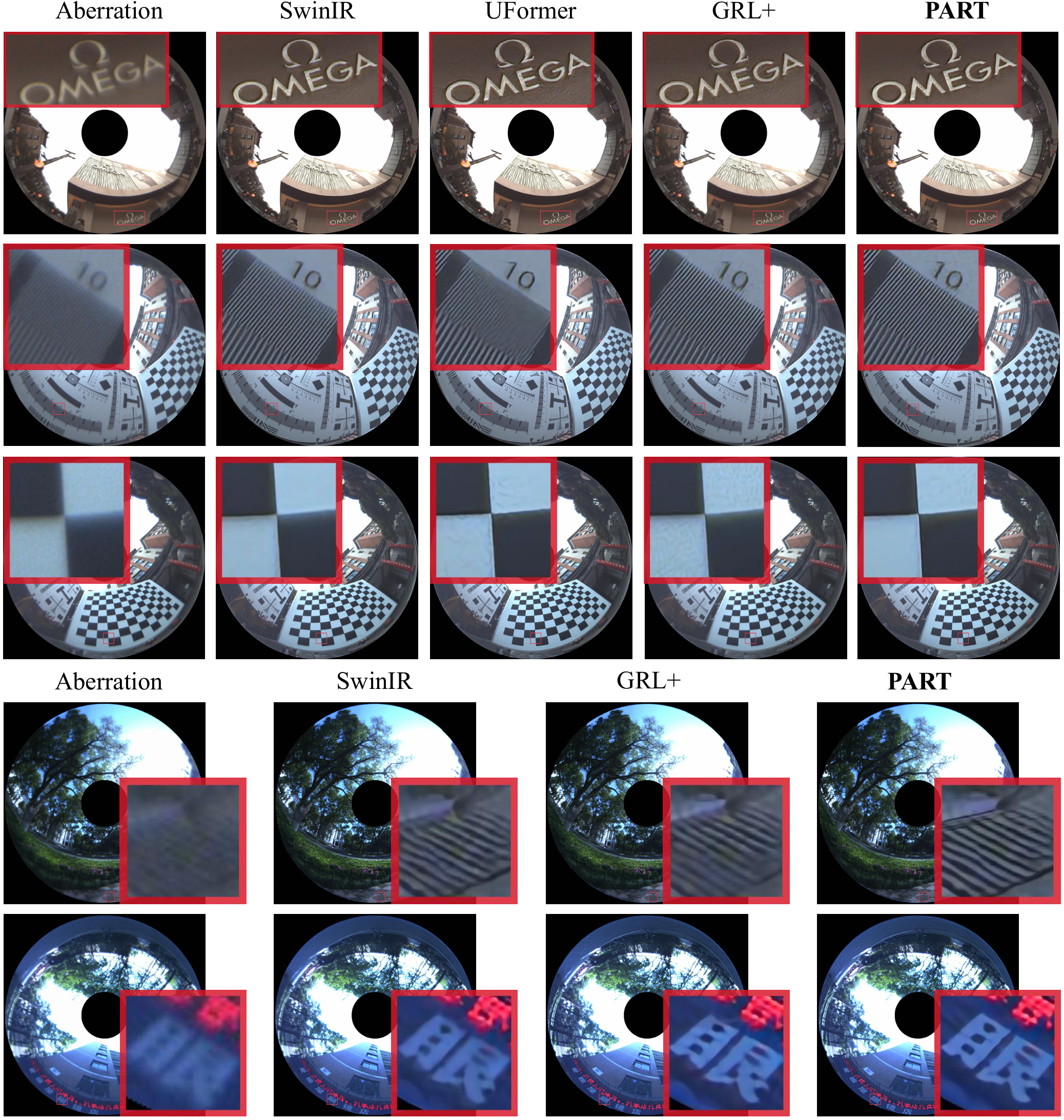}

  \caption{Visual results of PCIE on real MPIP with state-of-the-art models and our proposed PART. Top three rows: results of the AC pipeline. Bottom two rows: results of the SR$\&$AC pipeline. We choose the top-four performing methods to show the results, while UFormer is not applicable for SR$\&$AC.}
  \label{fig:real}

\end{figure}

\subsection{Experiments on Real-World Datasets}
\label{sec:real}
As shown in Table~\ref{tab:real}, PCIE with representative models makes significant contributions to the removal of the aberration-induced blur of real-world MPIP images. 
To be specific, OIQE improves from $55.22\%$ to $58.28\%{\sim}77.87\%$, and PSNR/SSM improves from $16.215dB/0.7995$ to $18.193dB/0.8690{\sim}19.841dB/0.8943$.
The results on NIQE and BRISQUE reveal a large variance, which is attributed to that these metrics are designed for perspective natural images rather than annular MPIP images.
For a comprehensive and intuitive evaluation, we rank each method on each metric and provide the average rank (A.R.).
In the real-world case, PART outperforms other models, achieving the best OIQE ($77.87\%$), and the best A.R. (2.2).
The subjective evaluation of the User Study (U.S.) also illustrates that PART delivers more visual-pleasant panoramic images, which has far superior selection rates. 
The visual results of PCIE on real-world scenes are provided in Fig.~\ref{fig:real}. 
PCIE enables most methods to deliver high-quality panoramic images with few aberrations and high resolution, where PART sets the state of the art in terms of higher contrast, sharper edges, and fewer artifacts.
Additionally, consistent with experiments on synthetic data, the recovered images of the SR$\&$AC pipeline reveal perceptually unpleasant artifacts.

\subsection{Investigation on GAN-based Training Strategies}
\label{sec:gan}
To generate richer details for recovered images, we investigate GAN-based training strategies on classical models RRDB, SwinIR, and our PART.
Following~\cite{liang2022details}, the GAN-based loss functions in ESRGAN~\cite{wang2018esrgan} and Local Discriminative Learning (LDL)~\cite{liang2022details} are adopted, where the former is a classical GAN-based framework and the latter is an improved strategy to remove artifacts. 
We take models trained with L1Loss, \ie~PSNR-oriented models, as pre-training generators, then apply GAN and LDL loss functions to enable these networks to generate more textures respectively.
As shown in Table~\ref{tab:gan}, on synthetic data, both GAN and LDL lead to a decrease in recovery accuracy (PSNR and SSIM), while bringing great gains under the perceptual quality metrics.
LDL is a more competitive strategy that outperforms GAN with higher fidelity and fewer visual artifacts, especially with PART.

\begin{figure}[!t]
  \centering
  \includegraphics[width=0.9\linewidth]{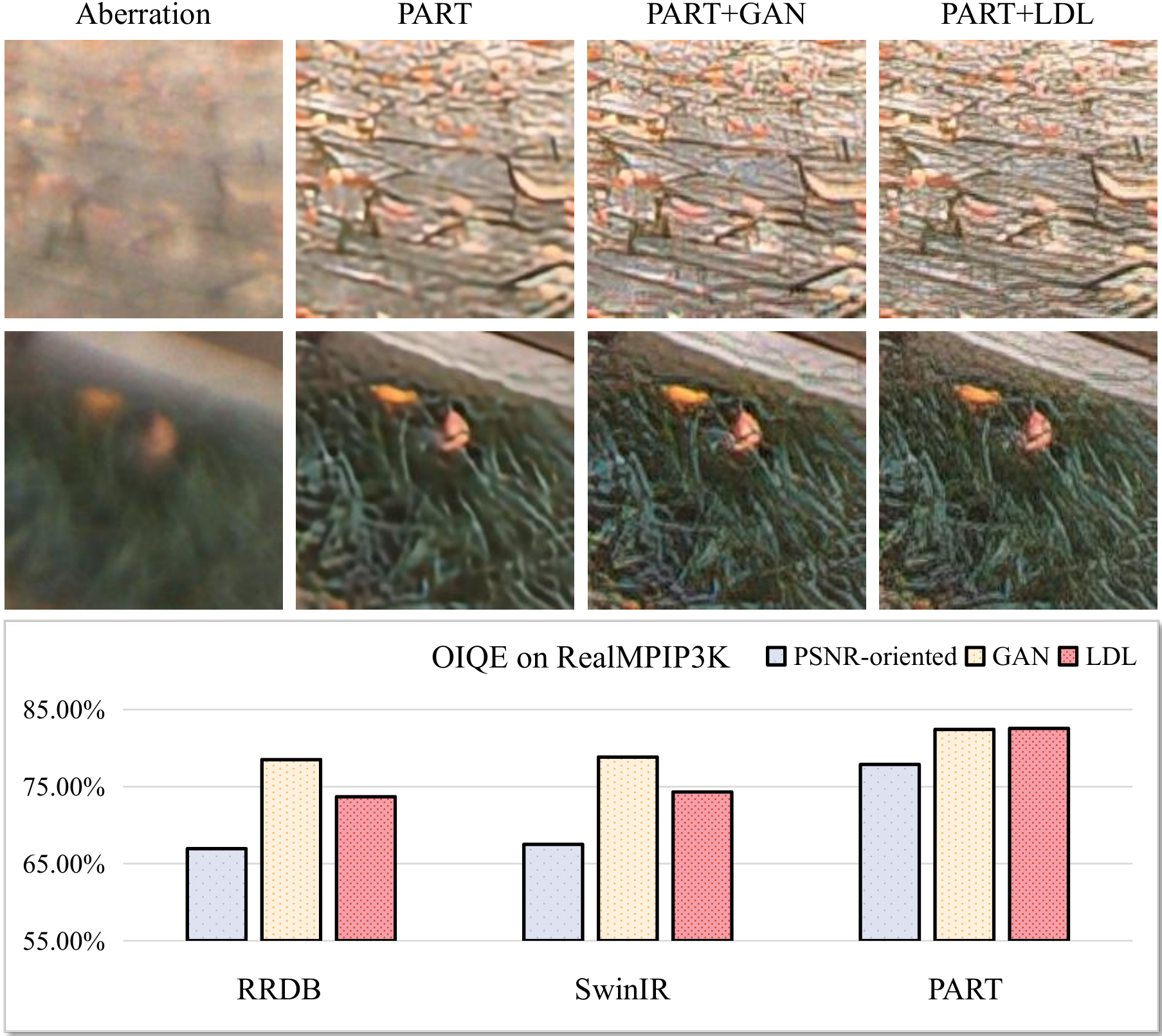}

  \caption{Evaluation of GAN-based training strategies on real-world data. We take the AC pipeline as an example, where image patches from RealMPIP3K-0031 and RealMPIP3K-0057 are presented.}
  \label{fig:gan}

\end{figure}

Regarding real-world data, we present the OIQE and qualitative results in Fig.~\ref{fig:gan}. 
GAN-based training further contributes to the removal of the aberration-induced blur, achieving better OIQE with higher image contrast.
Aside from this, GAN-based models deliver more realistic imaging results with richer textures, which also bring some perceptually unpleasant artifacts and fake details despite being trained with LDL.
We point out that the GAN-based strategies offer the potential for learning a more realistic high-quality MPIP image. Still, the local statistics in LDL of perspective images may need to be adapted to annular images for better suppression of artifacts.

We have further explored other potential generative models, \eg, the diffusion model~\cite{ho2020denoising,graikos2022diffusion}. Please refer to the Appendix for more results. 

\begin{table}[t]
    \begin{center}
        \caption{Quantitative evaluation of GAN-based training on our benchmarks of AC and SR$\&$AC.}

        \label{tab:gan}
        \input{Table/table_gan}
    \end{center}

\end{table}

\subsection{Effectiveness of PALHQ}
\label{sec:data}
The collected PALHQ demonstrates an impressive ability to train the model for recovering both synthetic and real-world MPIP images in previous experiments.
In this section, we explore whether PALHQ is necessary for PCIE. 
As an alternative to PALHQ, we simulate the aberrations of MPIP-P1 directly on the publicly available perspective image dataset, \ie~Flickr2K~\cite{timofte2017ntire}, creating PanoFlickr2K for training. 

\begin{figure}[!t]
  \centering
  \includegraphics[width=0.85\linewidth]{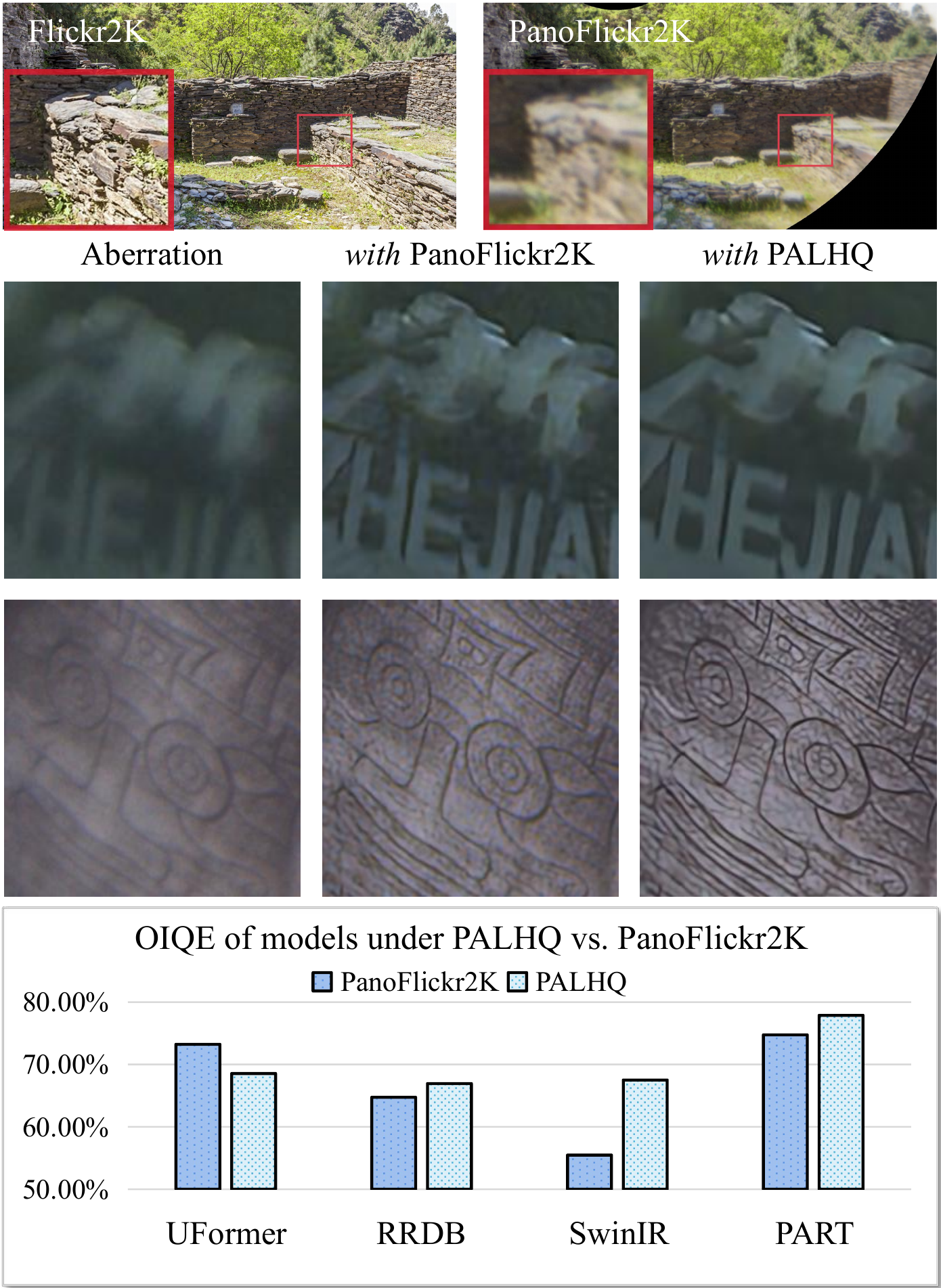}

  \caption{PALHQ \textit{vs.} PanoFlickr2K. The top row: illustration of PanoFlickr2K. We simulate the aberrations and FoV distributions of MPIP on perspective images of Flickr2K. Bottom two rows: image patches from RealMPIP3K-0020 and RealMPIP3K-0027. We take PART as an example to show the results trained on different datasets.}
  \label{fig:data}

\end{figure}

We compare representative models trained on PanoFlickr2K and PALHQ on both synthetic and real-world benchmarks in Table~\ref{tab:data} and Fig.~\ref{fig:data}.
It becomes clear that PALHQ contributes significantly to high-quality panoramic imaging, where the numerical results in all metrics are improved by a large margin and the visual results are more perceptually pleasant with sharper edges, fewer artifacts, and fewer noises. 

\begin{table}[t!]
    \begin{center}
        \caption{Quantitative comparison between the effectiveness of PALHQ and available HQ dataset in PCIE.}

        \label{tab:data}
        \input{Table/table_data}
    \end{center}

\end{table}

\subsection{Ablation Study}
\label{sec:ablation}
We conduct ablation studies to investigate how PSF-aware mechanisms contribute to high-quality MPIP image reconstruction. 
In all cases, the experiments are implemented with the AC pipeline on PALHQ-SynMPIP-P1, evaluated by PSNR and SSIM, and set up on the baseline model SwinIR. 

\noindent\textbf{Physical Information.}
As reported in Table~\ref{tab:pi}, the different types of physical information are concatenated with the input image respectively for an intuitive evaluation.
The PSF map contains rich information characterizing aberration-induced blur, providing better results compared to the FoV map. 
Then, we set the optimal $k'$ to $5$. 
A larger $k'$ tends to improve the model's scores, but the performance becomes saturated when $k'$ is too large with redundant and sparse information.

\noindent\textbf{PSF-aware Mechanisms.}
Table~\ref{tab:psf} shows that all the designed PSF-aware mechanisms contribute to reaching better scores, \ie~$0.230dB$ and $0.0013$ improvements in PSNR and SSIM. 
PFM attains the highest gains of $0.201dB$ in PSNR and $0.0012$ in SSIM. 
In addition, the performances of RRDB+ and GRL+ in Table~\ref{tab:synac} and Table~\ref{tab:synsrac} verify the consistent effectiveness of the plug-and-play PFM in other models, bringing improvements of $0.100dB{\sim}0.478dB$ in PSNR and $0.0006{\sim}0.0036$ in SSIM on SynMPIP-P1. 
Regarding the attention block, PMAB with $1{\times}1$ PFM and P-VSA enable adaptive self-attention guided by PSF information, outperforming the vanilla window-based self-attention. 

\noindent\textbf{Position of PFM.}
We further investigate the optimal position to insert PFM.
As shown in Table~\ref{tab:pfm}, we apply PFM after the feature extraction, at the last of each PRTB, and before the image reconstruction, for ablations. 
The PFM on shallow features reveals more competitive results, while increasing the number of PFM during the representation learning stage also leads to significant improvements. 
Using PFM in all ablated positions helps to reach the best performance, which is also corroborated by the observation in omnidirectional image super-resolution~\cite{yu2023osrt}.

\noindent\textbf{Effectiveness of PSF Representation.}
Table~\ref{tab:revised_ab} reports several possible PSF-aware mechanisms along with their vanilla versions without the guidance of the PSF map. The deformable (DConv and DeformSwin), FAC, and VSA mechanisms all deliver even worse performance compared to the baseline ($-0.957dB{\sim}-0.202dB$ in PSNR, $-0.0086{\sim}-0.0021$ in SSIM), which illustrates that the image-only network is unable to implicitly learn the complex spatial distribution of aberrations, leading to the unreliable predictions of offsets, convolution kernels, and varied-size windows.
Serving as a key modality, the PSF representation, \ie, the PSF map, which contains information of the intensity and size distribution of the PSF kernels, facilitates several potential PSF-aware mechanisms to achieve superior performance. To be specific, the guidance of PSF representation brings improvements of $0.261dB{\sim}1.158dB$ in PSNR and $0.0027{\sim}0.0098$ in SSIM to the vanilla mechanisms.

\begin{table*}[!t]

	\renewcommand\arraystretch{1.2}
	\setlength{\fboxrule}{0pt}
		\captionsetup{font={footnotesize}}

	\begin{center}
		\input{Table/ablation_all}
	\end{center}

\end{table*}

\begin{table}[h]
    \begin{center}
        \caption{Ablations on the effectiveness of PSF representation. Dconv: Deformable convolution~\cite{zhu2019deformable}, DeformSwin: Deformable Swin transformer~\cite{yu2023osrt}, ``P-'': the offsets are predicted from the PSF feature, FAC: Filter Adaptive Convolution~\cite{li2021involution, jiang2022fast}.}

        \label{tab:revised_ab}
        \input{Table/tab_revision_ablation}
    \end{center}

\end{table}

\subsection{Summary}
The extensive experiments illustrate the critical points in the proposed PCIE for achieving minimalist and high-quality panoramic imaging.
We summarize the following primary findings of our experiments:
\begin{compactitem}
    \item The proposed PCIE presents impressive high-quality imaging results, where the MPIP-P1 and AC pipeline are superior choices for delivering aberration-free panoramic images with much more realistic details.
    \item In PCIE, we find that window-attention-based models reveal better results. Furthermore, PSF-aware mechanisms are effective for improving the performance of SR models, where the proposed PSF-aware transformer, \ie~PART, sets state of the art.  
    \item The PSF representation plays a significant role in PSF-aware mechanisms, facilitating effective learning of the inverse process of the aberration-induced blur. 
    \item Regarding the training strategies, GAN-based methods contribute to more realistic recovered images, but with some visually unpleasant artifacts and fake details. The generative model appears to be more competitive in PCIE if a good balance is struck when generating rich details and suppressing artifacts.
    \item Comparing with the adaptation of perspective images, the collected high-quality panoramic annular images dataset, \ie~PALHQ, brings considerable improvements. PALHQ serves as the cornerstone of our PCIE for training a robust model to process MPIP images. 
\end{compactitem}

We hope that the PCIE can bring inspiration from optical design, network architecture, sensor choice,  data preparation, and training strategies, for minimalist and high-quality panoramic imaging in mobile and wearable applications.

%% file: Table/synall.tex
\begin{minipage}[ht]{0.63\textwidth}
	\begin{center}
		\captionsetup{font={scriptsize}}
	    \caption{Quantitative evaluation of PCIE with the AC pipeline on synthetic benchmarks with MPIP-P1 and MPIP-P2. We highlight the \YKL{best} and \JQ{second} results. The ``*'' for NAFNet and Restormer denotes that the cropping testing strategy is applied. }

		\setlength{\tabcolsep}{1mm}
		\resizebox{\textwidth}{!}{
                \begin{tabular}{|cc|cccc|cccc|}
                \hline
                \multicolumn{2}{|c|}{\multirow{2}{*}{Method}}                    & \multicolumn{4}{c|}{PALHQ-SynMPIP-P1}                                                    & \multicolumn{4}{c|}{PALHQ-SynMPIP-P2}                                                    \\ \cline{3-10} 
                \multicolumn{2}{|c|}{}                                           & \multicolumn{1}{c|}{PSNR$\uparrow$} & \multicolumn{1}{c|}{SSIM$\uparrow$} & \multicolumn{1}{c|}{LPIPS$\downarrow$} & FID$\downarrow$ & \multicolumn{1}{c|}{PSNR$\uparrow$} & \multicolumn{1}{c|}{SSIM$\uparrow$} & \multicolumn{1}{c|}{LPIPS$\downarrow$} & FID$\downarrow$ \\ \hline \hline
                \multicolumn{1}{|c|}{\multirow{7}{*}{SR}}     & RRDB~\cite{wang2018esrgan}           & \multicolumn{1}{c|}{32.716}     & \multicolumn{1}{c|}{0.9265}     & \multicolumn{1}{c|}{0.0469}      &03.704     & \multicolumn{1}{c|}{26.996}     & \multicolumn{1}{c|}{0.8503}     & \multicolumn{1}{c|}{0.0896}      & 17.413    \\
                \multicolumn{1}{|c|}{}                        & RCAN~\cite{zhang2018image}            & \multicolumn{1}{c|}{32.496}     & \multicolumn{1}{c|}{0.9257}     & \multicolumn{1}{c|}{0.0459}      &04.327     & \multicolumn{1}{c|}{26.456}     & \multicolumn{1}{c|}{0.8443}     & \multicolumn{1}{c|}{0.0956}      & 22.619    \\
                \multicolumn{1}{|c|}{}                        & EDSR~\cite{lim2017enhanced}            & \multicolumn{1}{c|}{32.868}     & \multicolumn{1}{c|}{0.9282}     & \multicolumn{1}{c|}{0.0449}      &03.770     & \multicolumn{1}{c|}{26.951}     & \multicolumn{1}{c|}{0.8500}     & \multicolumn{1}{c|}{0.0889}      &17.871     \\
                \multicolumn{1}{|c|}{}                        & SwinIR~\cite{liang2021swinir}           & \multicolumn{1}{c|}{32.913}     & \multicolumn{1}{c|}{0.9291}     & \multicolumn{1}{c|}{0.0446}      &   03.670  & \multicolumn{1}{c|}{26.935}     & \multicolumn{1}{c|}{0.8509}     & \multicolumn{1}{c|}{0.0884}      &   17.630  \\
                \multicolumn{1}{|c|}{}                        & EDT~\cite{li2021efficient}              & \multicolumn{1}{c|}{32.929}     & \multicolumn{1}{c|}{0.9288}     & \multicolumn{1}{c|}{0.0450}      &03.658    & \multicolumn{1}{c|}{27.055}     & \multicolumn{1}{c|}{0.8518}     & \multicolumn{1}{c|}{0.0888}      &16.976     \\
                \multicolumn{1}{|c|}{}                        & HAT~\cite{chen2205activating}              & \multicolumn{1}{c|}{32.925}     & \multicolumn{1}{c|}{0.9288}     & \multicolumn{1}{c|}{0.0447}      &03.748     & \multicolumn{1}{c|}{26.827}     & \multicolumn{1}{c|}{0.8500}     & \multicolumn{1}{c|}{0.0889}      &18.498     \\
                \multicolumn{1}{|c|}{}                        & GRL~\cite{li2023efficient}              & \multicolumn{1}{c|}{32.369}     & \multicolumn{1}{c|}{0.9256}     & \multicolumn{1}{c|}{0.0457}      &04.234     & \multicolumn{1}{c|}{26.268}     & \multicolumn{1}{c|}{0.8424}     & \multicolumn{1}{c|}{0.0943}      &22.463     \\ \hline \hline
                \multicolumn{1}{|c|}{\multirow{4}{*}{Deblur}} & HINet~\cite{chen2021hinet}            & \multicolumn{1}{c|}{32.238}     & \multicolumn{1}{c|}{0.9234}     & \multicolumn{1}{c|}{0.0476}      &04.159     & \multicolumn{1}{c|}{26.401}     & \multicolumn{1}{c|}{0.8428}     & \multicolumn{1}{c|}{0.0933}      & 22.341    \\
                \multicolumn{1}{|c|}{}                        & NAFNet*~\cite{chen2022simple}           & \multicolumn{1}{c|}{32.837}     & \multicolumn{1}{c|}{0.9274}     & \multicolumn{1}{c|}{\JQ{0.0441}}      &03.845     & \multicolumn{1}{c|}{27.045}     & \multicolumn{1}{c|}{0.8514}     & \multicolumn{1}{c|}{0.0856}      & 17.504     \\
                \multicolumn{1}{|c|}{}                        & Restormer*~\cite{zamir2022restormer}        & \multicolumn{1}{c|}{32.971}     & \multicolumn{1}{c|}{0.9287}     & \multicolumn{1}{c|}{0.0445}      &03.763     & \multicolumn{1}{c|}{27.001}     & \multicolumn{1}{c|}{0.8510}     & \multicolumn{1}{c|}{0.0870}      &17.023     \\
                \multicolumn{1}{|c|}{}                        & Uformer~\cite{wang2022uformer}          & \multicolumn{1}{c|}{\JQ{32.999}}     & \multicolumn{1}{c|}{0.9290}     & \multicolumn{1}{c|}{{0.0442}}      &03.672     & \multicolumn{1}{c|}{\JQ{27.133}}     & \multicolumn{1}{c|}{0.8525}     & \multicolumn{1}{c|}{0.0866}      & 16.693    \\ \hline \hline
                \multicolumn{1}{|c|}{\multirow{4}{*}{PSF-aware}}     & PI$^2$RNet~\cite{jiang2023annular}          & \multicolumn{1}{c|}{32.682}     & \multicolumn{1}{c|}{0.9268}     & \multicolumn{1}{c|}{0.0448}      &03.638     & \multicolumn{1}{c|}{26.656}     & \multicolumn{1}{c|}{0.8471}     & \multicolumn{1}{c|}{0.0874}      &18.544     \\
                \multicolumn{1}{|c|}{}                        
                & \textbf{RRDB+}          & \multicolumn{1}{c|}{32.816}     & \multicolumn{1}{c|}{0.9271}     & \multicolumn{1}{c|}{0.0456}      &03.746     & \multicolumn{1}{c|}{27.050}     & \multicolumn{1}{c|}{0.8505}     & \multicolumn{1}{c|}{0.0895}      & 17.103    \\
                \multicolumn{1}{|c|}{}                        
                                                                & \textbf{GRL+}             & \multicolumn{1}{c|}{32.847}     & \multicolumn{1}{c|}{\JQ{0.9292}}     & \multicolumn{1}{c|}{0.0454}      & \JQ{03.627}    & \multicolumn{1}{c|}{27.020}     & \multicolumn{1}{c|}{\JQ{0.8528}}     & \multicolumn{1}{c|}{\JQ{0.0864}}      &\YKL{16.281}     \\
                \multicolumn{1}{|c|}{}                        & \textbf{PART (Ours)} & \multicolumn{1}{c|}{\YKL{33.143}}     & \multicolumn{1}{c|}{\YKL{0.9304}}     & \multicolumn{1}{c|}{\YKL{0.0435}}      & \YKL{03.571}    & \multicolumn{1}{c|}{\YKL{27.198}}     & \multicolumn{1}{c|}{\YKL{0.8540}}     & \multicolumn{1}{c|}{\YKL{0.0855}}      &\JQ{16.436}     \\ \hline
                \end{tabular}
		}
	    \label{tab:synac}
	\end{center}
\end{minipage}
\hspace{2pt}
\begin{minipage}[ht]{0.33\textwidth}
	\begin{center}
		\captionsetup{font={scriptsize}}
		\caption{Quantitative evaluation of PCIE with SR$\&$AC pipeline on synthetic benchmark.}

		\setlength{\tabcolsep}{1mm}
		\resizebox{\textwidth}{!}{
                \begin{tabular}{|c|cccc|}
                \hline
                \multirow{2}{*}{Method} & \multicolumn{4}{c|}{PALHQ-SynMPIP-P1}                                                    \\ \cline{2-5} 
                                        & \multicolumn{1}{c|}{PSNR$\uparrow$} & \multicolumn{1}{c|}{SSIM$\uparrow$} & \multicolumn{1}{c|}{LPIPS$\downarrow$} & FID$\downarrow$ \\ \hline \hline
                RRDB~\cite{wang2018esrgan}                    & \multicolumn{1}{c|}{28.856}     & \multicolumn{1}{c|}{0.8758}     & \multicolumn{1}{c|}{0.0733}      & 09.957    \\
                RCAN~\cite{zhang2018image}                    & \multicolumn{1}{c|}{28.238}     & \multicolumn{1}{c|}{0.8686}     & \multicolumn{1}{c|}{0.0787}      & 12.689     \\
                EDSR~\cite{lim2017enhanced}                    & \multicolumn{1}{c|}{28.817}     & \multicolumn{1}{c|}{0.8759}     & \multicolumn{1}{c|}{0.0715}      & 10.670    \\
                SwinIR~\cite{liang2021swinir}                  & \multicolumn{1}{c|}{28.985}     & \multicolumn{1}{c|}{\JQ{0.8781}}     & \multicolumn{1}{c|}{\JQ{0.0714}}      & 09.938    \\
                EDT~\cite{li2021efficient}                     & \multicolumn{1}{c|}{29.008}     & \multicolumn{1}{c|}{0.8777}     & \multicolumn{1}{c|}{0.0726}      & 10.750    \\
                HAT~\cite{chen2205activating}                     & \multicolumn{1}{c|}{28.921}     & \multicolumn{1}{c|}{0.8771}     & \multicolumn{1}{c|}{0.0727}      & 10.141     \\
                GRL~\cite{li2023efficient}                     & \multicolumn{1}{c|}{28.695}     & \multicolumn{1}{c|}{0.8753}     & \multicolumn{1}{c|}{\JQ{0.0714}}      &09.829     \\\hline \hline
                \textbf{RRDB+}                 & \multicolumn{1}{c|}{\JQ{29.044}}     & \multicolumn{1}{c|}{0.8774}     & \multicolumn{1}{c|}{0.0724}      &10.068     \\
                \textbf{GRL+ }                   & \multicolumn{1}{c|}{28.757}     & \multicolumn{1}{c|}{0.8768}     & \multicolumn{1}{c|}{0.0716}      & \JQ{09.709}    \\
                \textbf{PART (Ours)}     & \multicolumn{1}{c|}{\YKL{29.310}}     & \multicolumn{1}{c|}{\YKL{0.8819}}     & \multicolumn{1}{c|}{\YKL{0.0681}}      & \YKL{09.648}     \\ \hline
                \end{tabular}
		}
		\label{tab:synsrac}
	\end{center}
\end{minipage}

%% file: Table/table_real_revised.tex
\resizebox{0.85\textwidth}{!}
{
\renewcommand{\arraystretch}{1.2}
\setlength{\tabcolsep}{1mm}{

\begin{tabular}{|c|ccc|ccc|ccc|c|}
\hline
\multirow{2}{*}{Method} & \multicolumn{3}{c|}{RealMPIP3K-Checkerboard} & \multicolumn{3}{c|}{RealMPIP3K-AC} & \multicolumn{3}{c|}{RealMPIP3K-SR\&AC} & \multirow{2}{*}{A.R.$\downarrow$} \\ \cline{2-10}
 & \multicolumn{1}{c|}{OIQE$\uparrow$} & \multicolumn{1}{c|}{PSNR$\uparrow$} & SSIM$\uparrow$ & \multicolumn{1}{c|}{U.S.$\uparrow$} & \multicolumn{1}{c|}{NIQE$\downarrow$} & BRISQUE$\downarrow$ & \multicolumn{1}{c|}{U.S.$\uparrow$} & \multicolumn{1}{c|}{NIQE$\downarrow$} & BRISQUE$\downarrow$ &  \\ \hline\hline
RRDB~\cite{wang2018esrgan} & \multicolumn{1}{c|}{66.94\%(4)} & \multicolumn{1}{c|}{19.587(3)} & 0.8872(4) & \multicolumn{1}{c|}{51.91\%(3)} & \multicolumn{1}{c|}{04.930(8)} & 45.692(2) & \multicolumn{1}{c|}{37.84\%(5)} & \multicolumn{1}{c|}{04.848(4)} & 50.729(5) & 4.2 \\
SwinIR~\cite{liang2021swinir} & \multicolumn{1}{c|}{67.51\%(3)} & \multicolumn{1}{c|}{18.991(5)} & 0.8863(5) & \multicolumn{1}{c|}{49.05\%(4)} & \multicolumn{1}{c|}{04.816(6)} & 46.380(5) & \multicolumn{1}{c|}{38.92\%(4)} & \multicolumn{1}{c|}{04.833(2)} & 50.555(4) & 4.2 \\
GRL~\cite{li2023efficient} & \multicolumn{1}{c|}{64.63\%(7)} & \multicolumn{1}{c|}{19.348(4)} & 0.8885(3) & \multicolumn{1}{c|}{36.43\%(7)} & \multicolumn{1}{c|}{04.665(1)} & 46.587(6) & \multicolumn{1}{c|}{08.65\%(6)} & \multicolumn{1}{c|}{04.824(1)} & 51.057(6) & 4.6 \\ \hline\hline
UFormer~\cite{wang2022uformer} & \multicolumn{1}{c|}{68.55\%(2)} & \multicolumn{1}{c|}{19.841(1)} & 0.8897(2) & \multicolumn{1}{c|}{61.43\%(2)} & \multicolumn{1}{c|}{04.914(7)} & 45.427(1) & \multicolumn{1}{c|}{\textit{n.a.}} & \multicolumn{1}{c|}{\textit{n.a.}} & \textit{n.a.} & \JQ{2.5} \\ \hline\hline
PI$^2$RNet~\cite{jiang2023annular} & \multicolumn{1}{c|}{64.86\%(6)} & \multicolumn{1}{c|}{18.443(7)} & 0.8734(7) & \multicolumn{1}{c|}{42.14\%(6)} & \multicolumn{1}{c|}{04.710(3)} & 47.148(8) & \multicolumn{1}{c|}{\textit{n.a.}} & \multicolumn{1}{c|}{\textit{n.a.}} & \textit{n.a.} & 6.2 \\
\textbf{RRDB+} & \multicolumn{1}{c|}{58.28\%(8)} & \multicolumn{1}{c|}{18.458(6)} & 0.8758(6) & \multicolumn{1}{c|}{32.14\%(8)} & \multicolumn{1}{c|}{04.783(5)} & 46.675(7) & \multicolumn{1}{c|}{62.16\%(3)} & \multicolumn{1}{c|}{04.857(5)} & 50.383(2) & 5.6 \\
\textbf{GRL+} & \multicolumn{1}{c|}{65.31\%(5)} & \multicolumn{1}{c|}{18.193(8)} & 0.8690(8) & \multicolumn{1}{c|}{48.33\%(5)} & \multicolumn{1}{c|}{04.724(4)} & 46.138(4) & \multicolumn{1}{c|}{68.92\%(2)} & \multicolumn{1}{c|}{04.842(3)} & 50.061(1) & 4.4 \\
\textbf{PART} & \multicolumn{1}{c|}{77.87\%(1)} & \multicolumn{1}{c|}{19.606(2)} & 0.8943(1) & \multicolumn{1}{c|}{78.57\%(1)} & \multicolumn{1}{c|}{04.707(2)} & 45.968(3) & \multicolumn{1}{c|}{83.51\%(1)} & \multicolumn{1}{c|}{04.933(6)} & 50.422(3) & \YKL{2.2} \\ \hline
\end{tabular}

}
}

%% file: Table/table_gan.tex
\resizebox{0.5\textwidth}{!}
{
\renewcommand{\arraystretch}{1.2}
\setlength{\tabcolsep}{1mm}{
\begin{tabular}{|c|c|c|cccc|ccc|}
\hline
\multirow{2}{*}{Task}   & \multirow{2}{*}{Method} & \multirow{2}{*}{Training Strategy} & \multicolumn{4}{c|}{PALHQ-SynMPIP-P1}                                                                                                                                      \\ \cline{4-7} 
                        &                         &                                    & \multicolumn{1}{c|}{PSNR$\uparrow$} & \multicolumn{1}{c|}{SSIM$\uparrow$} & \multicolumn{1}{c|}{LPIPS$\downarrow$} & FID$\downarrow$                \\ \hline \hline
\multirow{9}{*}{AC}     & \multirow{3}{*}{RRDB~\cite{wang2018esrgan}}   & PSNR-oriented                      & \multicolumn{1}{c|}{32.716}     & \multicolumn{1}{c|}{0.9265}     & \multicolumn{1}{c|}{0.0469}      & 03.704            \\
                        &                         & +GAN~\cite{wang2018esrgan}                               & \multicolumn{1}{c|}{28.929}     & \multicolumn{1}{c|}{0.8840}     & \multicolumn{1}{c|}{0.0392}      & 04.919                                   \\
                        &                         & +LDL~\cite{liang2022details}                               & \multicolumn{1}{c|}{31.864}     & \multicolumn{1}{c|}{0.9166}     & \multicolumn{1}{c|}{0.0338}      & 04.559           \\
                        & \multirow{3}{*}{SwinIR~\cite{liang2021swinir}} & PSNR-oriented                      & \multicolumn{1}{c|}{\JQ{32.913}}     & \multicolumn{1}{c|}{\JQ{0.9291}}     & \multicolumn{1}{c|}{0.0446}      & 03.670                               \\
                        &                         & +GAN~\cite{wang2018esrgan}                              & \multicolumn{1}{c|}{29.916}     & \multicolumn{1}{c|}{0.8920}     & \multicolumn{1}{c|}{0.0449}      &04.254                               \\
                        &                         & +LDL~\cite{liang2022details}                               & \multicolumn{1}{c|}{31.770}     & \multicolumn{1}{c|}{0.9130}     & \multicolumn{1}{c|}{\JQ{0.0297}}      &\YKL{03.444}                               \\
                        & \multirow{3}{*}{\textbf{PART}}   & PSNR-oriented                      & \multicolumn{1}{c|}{\YKL{33.143}}     & \multicolumn{1}{c|}{\YKL{0.9304}}     & \multicolumn{1}{c|}{0.0435}      & 03.571                                  \\
                        &                         & +GAN~\cite{wang2018esrgan}                               & \multicolumn{1}{c|}{30.965}     & \multicolumn{1}{c|}{0.9045}     & \multicolumn{1}{c|}{0.0410}      &04.402                                    \\
                        &                         & +LDL~\cite{liang2022details}                               & \multicolumn{1}{c|}{31.854}     & \multicolumn{1}{c|}{0.9148}     & \multicolumn{1}{c|}{\YKL{0.0264}}      &\JQ{03.541}                                 \\ \hline \hline
\multirow{9}{*}{SR\&AC} & \multirow{3}{*}{RRDB~\cite{wang2018esrgan}}   & PSNR-oriented                      & \multicolumn{1}{c|}{28.856}     & \multicolumn{1}{c|}{0.8758}     & \multicolumn{1}{c|}{0.0733}      &09.957             \\
                        &                         & +GAN~\cite{wang2018esrgan}                               & \multicolumn{1}{c|}{25.842}     & \multicolumn{1}{c|}{0.8276}     & \multicolumn{1}{c|}{0.0638}      & 12.564           \\
                        &                         & +LDL~\cite{liang2022details}                               & \multicolumn{1}{c|}{28.112}     & \multicolumn{1}{c|}{0.8633}     & \multicolumn{1}{c|}{\JQ{0.0561}}      &09.746                                  \\
                        & \multirow{3}{*}{SwinIR~\cite{liang2021swinir}} & PSNR-oriented                      & \multicolumn{1}{c|}{\JQ{28.985}}     & \multicolumn{1}{c|}{\JQ{0.8781}}     & \multicolumn{1}{c|}{0.0714}      &09.938                                 \\
                        &                         & +GAN~\cite{wang2018esrgan}                               & \multicolumn{1}{c|}{26.596}     & \multicolumn{1}{c|}{0.8385}     & \multicolumn{1}{c|}{0.0634}      &12.155          \\
                        &                         & +LDL~\cite{liang2022details}                               & \multicolumn{1}{c|}{27.875}     & \multicolumn{1}{c|}{0.8575}     & \multicolumn{1}{c|}{0.0686}      &09.423                              \\
                        & \multirow{3}{*}{\textbf{PART}}   & PSNR-oriented                      & \multicolumn{1}{c|}{\YKL{29.310}}     & \multicolumn{1}{c|}{\YKL{0.8819}}     & \multicolumn{1}{c|}{0.0681}      & 09.648                            \\
                        &                         & +GAN~\cite{wang2018esrgan}                               & \multicolumn{1}{c|}{28.382}     & \multicolumn{1}{c|}{0.8688}     & \multicolumn{1}{c|}{0.0682}      &\JQ{08.897}                                \\
                        &                         & +LDL~\cite{liang2022details}                               & \multicolumn{1}{c|}{28.608}     & \multicolumn{1}{c|}{0.8720}     & \multicolumn{1}{c|}{\YKL{0.0508}}      &\YKL{08.715}           \\ \hline
\end{tabular}

}
}

%% file: Table/table_data.tex
\resizebox{0.5\textwidth}{!}
{
\renewcommand{\arraystretch}{1.2}
\setlength{\tabcolsep}{1mm}{
\begin{tabular}{|c|c|c|cccc|}
\hline
\multirow{2}{*}{Task}   & \multirow{2}{*}{Method} & \multirow{2}{*}{Training Dataset} & \multicolumn{4}{c|}{PALHQ-SynMPIP-P1}                                                                                                                         \\ \cline{4-7} 
                        &                         &                                   & \multicolumn{1}{c|}{PSNR$\uparrow$} & \multicolumn{1}{c|}{SSIM$\uparrow$} & \multicolumn{1}{c|}{LPIPS$\downarrow$} & FID$\downarrow$ \\ \hline \hline
\multirow{8}{*}{AC}     & \multirow{2}{*}{Uformer~\cite{wang2022uformer}} & PanoFlickr2K                      & \multicolumn{1}{c|}{32.340}     & \multicolumn{1}{c|}{0.8253}     & \multicolumn{1}{c|}{0.0484}      &04.558          \\
                        &                         & \textbf{PALHQ}                             & \multicolumn{1}{c|}{32.999}     & \multicolumn{1}{c|}{0.9290}     & \multicolumn{1}{c|}{0.0442}      &03.672         \\
                        & \multirow{2}{*}{RRDB~\cite{wang2018esrgan}}   & PanoFlickr2K                      & \multicolumn{1}{c|}{32.128}     & \multicolumn{1}{c|}{0.9235}     & \multicolumn{1}{c|}{0.0508}      &04.717          \\
                        &                         & \textbf{PALHQ}                             & \multicolumn{1}{c|}{32.716}     & \multicolumn{1}{c|}{0.9265}     & \multicolumn{1}{c|}{0.0469}      & 03.704         \\
                        & \multirow{2}{*}{SwinIR~\cite{liang2021swinir}} & PanoFlickr2K                      & \multicolumn{1}{c|}{32.292}     & \multicolumn{1}{c|}{0.9255}     & \multicolumn{1}{c|}{0.0485}      &04.521         \\
                        &                         & \textbf{PALHQ}                             & \multicolumn{1}{c|}{32.913}     & \multicolumn{1}{c|}{0.9291}     & \multicolumn{1}{c|}{0.0446}      & 03.670          \\
                        & \multirow{2}{*}{PART}   & PanoFlickr2K                      & \multicolumn{1}{c|}{32.498}     & \multicolumn{1}{c|}{0.9259}     & \multicolumn{1}{c|}{0.0480}      &04.348         \\
                        &                         & \textbf{PALHQ}                             & \multicolumn{1}{c|}{33.143}     & \multicolumn{1}{c|}{0.9304}     & \multicolumn{1}{c|}{0.0435}      & 03.571       \\ \hline \hline
\multirow{6}{*}{SR\&AC} & \multirow{2}{*}{RRDB~\cite{wang2018esrgan}}   & PanoFlickr2K                      & \multicolumn{1}{c|}{27.943}     & \multicolumn{1}{c|}{0.8688}     & \multicolumn{1}{c|}{0.0906}      & 12.276         \\
                        &                         & \textbf{PALHQ}                             & \multicolumn{1}{c|}{28.856}     & \multicolumn{1}{c|}{0.8758}     & \multicolumn{1}{c|}{0.0733}      & 09.957       \\
                        & \multirow{2}{*}{SwinIR~\cite{liang2021swinir}} & PanoFlickr2K                      & \multicolumn{1}{c|}{28.129}     & \multicolumn{1}{c|}{0.8705}     & \multicolumn{1}{c|}{0.0880}      &12.064         \\
                        &                         & \textbf{PALHQ}                             & \multicolumn{1}{c|}{28.985}     & \multicolumn{1}{c|}{0.8781}     & \multicolumn{1}{c|}{0.0714}      &09.938         \\
                        & \multirow{2}{*}{PART}   & PanoFlickr2K                      & \multicolumn{1}{c|}{28.558}     & \multicolumn{1}{c|}{0.8748}     & \multicolumn{1}{c|}{0.0854}      & 12.078        \\
                        &                         & \textbf{PALHQ}                             & \multicolumn{1}{c|}{29.310}     & \multicolumn{1}{c|}{0.8819}     & \multicolumn{1}{c|}{0.0681}      &09.648      \\ \hline
\end{tabular}

}
}

%% file: Table/ablation_all.tex
\begin{minipage}[ht]{0.3\textwidth}
	\begin{center}
		\captionsetup{font={scriptsize}}
	    \caption{Ablations on Physical Information.}

		\setlength{\tabcolsep}{2mm}
		\resizebox{\textwidth}{!}{
                \begin{tabular}{cc|cc}
                \hline 
                Physical Information     & \textit{k'} & PSNR & SSIM  \\ \hline \hline
                \textit{w/o}             & -          &32.913      &0.9291            \\
                FoV map                  & -          &32.992      &0.9293             \\ \hline \hline
                 & 1          &33.012      &0.9300             \\
                 \rowcolor{gray!15}{PSF map} & 5          &\textbf{33.021}      &\textbf{0.9301}             \\
                                         & 9          &\textbf{33.021}      &0.9299             \\ \hline
                \end{tabular}
		}
	    \label{tab:pi}
	\end{center}
\end{minipage}
\hspace{2pt}
\begin{minipage}[ht]{0.35\textwidth}
	\begin{center}
		\captionsetup{font={scriptsize}}
		\caption{Ablations on PSF-aware Mechanisms.}

		\setlength{\tabcolsep}{2mm}
		\resizebox{\textwidth}{!}{
                \begin{tabular}{cc|ccc}
                \hline
                \multicolumn{2}{c|}{PSF-aware Mechanism} & Params                & PSNR                 & SSIM                                 \\ \hline \hline
                \textit{w/o}           & -               &11.97M                      & 32.913                     & 0.9291                                         \\
                concat                 & -               &12.02M                      & 33.021                     & 0.9301                                       \\
                PFM                    & -               &16.72M                      & 33.114                     & 0.9303                                          \\
                \multirow{3}{*}{PMAB}   & 1$\times$1 PFM       &14.32M                      & 33.069                     & 0.9302                                         \\
                                       & P-VSA            & 12.18M                     & 32.999                      & 0.9297   
                                                    \\
                                       & \textit{both}   &14.53M                      & 33.082                     &  0.9303                                        \\
                \rowcolor{gray!15}\textit{all}           & -               & \multicolumn{1}{l}{19.27M} & \multicolumn{1}{l}{\textbf{33.143}} & \multicolumn{1}{l}{\textbf{0.9304}}  \\ \hline
                \end{tabular}
		}
		\label{tab:psf}
	\end{center}
\end{minipage}
\hspace{2pt}
\begin{minipage}[ht]{0.24\textwidth}
	\begin{center}
		\captionsetup{font={scriptsize}}
	    \caption{Ablations on the Position of PFM.}

		\setlength{\tabcolsep}{1.0mm}
		\resizebox{\textwidth}{!}{
                \begin{tabular}{c|lccc}
                \hline
                Position     & Params & PSNR                 & SSIM                                 \\ \hline \hline
                \textit{w/o} &11.97M        & 32.913                     &  0.9291                                        \\
                first conv   &12.61M        & 33.032                     &  0.9297                                         \\
                PRTB         &15.54M        & 33.071                     &  0.9299                                         \\
                last conv    &12.61M        & 32.971                     &  0.9298                                        \\
                \rowcolor{gray!15}\textit{all} &16.72M        & \multicolumn{1}{l}{\textbf{33.114}} & \multicolumn{1}{l}{\textbf{0.9303}}  \\ \hline
                \end{tabular}
		}
	    \label{tab:pfm}
	\end{center}
\end{minipage}
\hspace{5pt}

%% file: Table/tab_revision_ablation.tex
\resizebox{0.35\textwidth}{!}
{
\renewcommand{\arraystretch}{1.2}
\setlength{\tabcolsep}{1mm}{

\begin{tabular}{c|ccc}
\hline
Method         & Params & PSNR & SSIM \\ \hline \hline
baseline       & 11.97M & 32.913 & 0.9291      \\\hline
\textit{w} Dconv        &14.71M &32.258 &0.9237      \\
\rowcolor{gray!15}\textit{w} P-Dconv      &14.71M &33.064 &0.9303      \\\hline
\textit{w} FAC          &16.72M &31.956 &0.9205      \\
\rowcolor{gray!15}\textit{w} PFM          &16.72M &33.114 &0.9303     \\  \hline
\textit{w} DeformSwin   &13.21M &32.711 &0.9270\\
\rowcolor{gray!15}\textit{w} P-DeformSwin &13.21M &32.972 &0.9297       \\\hline
\textit{w} VSA          &12.18M &32.496 &0.9253     \\
\rowcolor{gray!15}\textit{w} P-VSA        &12.18M &32.999 &0.9297      \\ \hline
\end{tabular}

}
}

%% file: Tex_content/conclusion_revised.tex
\subsection{Conclusion}
In this paper, we design PCIE to present a general solution to minimalist and high-quality panoramic imaging. 
Based on the idea of PAL, the MPIP is proposed for 360{\textdegree} panoramic imaging with less than three lenses. 
Then, learning-based models, which are trained on synthetic aberration images from simulation, are applied to solve the aberration-induced blur and low resolution of MPIP images.
A new dataset PALHQ is collected to fill the gap of high-quality PAL images for low-level vision. 
We explore utilizing PSF information of the optical system to improve the performance of models and design a PSF-aware transformer PART with PSF-aware mechanisms.
The plug-and-play mechanism PFM can enhance modern SR models for removing aberration-induced blur, while PART with PMAB delivers state-of-the-art performance on both synthetic and real-world benchmarks. 
Extensive experiments are conducted to investigate how to improve PCIE, providing heuristic findings for constructing a computational-imaging-based minimalist panoramic system with impressive imaging quality, in terms of optical design, network architecture, sensor selection, training strategies, and data preparation.  

\subsection{Discussion and Future Work}
There are still some limitations in PCIE, which call for further investigation into extremely high-quality imaging. 
First, the PSF-aware mechanisms are designed in a straightforward way, which improves the performance, yet, with extra parameters and computational overhead.
More efficient and effective PSF-aware architectures or training strategies are expected to further enhance the performance.
Meanwhile, the improvements on CNN-based models are less pronounced compared to those on transformer models.
We are interested in the design of learnable PSF representation, PSF-aware dynamic, deformable, and dilated convolution, or PSF-aware varied-shape window attention for better exploration of PSF information.
Then, we investigate state-of-the-art GAN-based training strategies, while there is open research space for further suppressing artifacts.
Aside from this, the results of PCIE on real-world data are not as good as on synthetic data, where artifacts and fake details exist in some recovered images. 
The considerable synthetic-to-real gap needs future research on domain adaptation. 
The image number of PALHQ is also limited due to the difficulties of capturing high-quality PAL images under various scenes. 
We intend to design a hybrid training approach to take advantage of the large data size of the publicly available perspective datasets while improving the training with PALHQ.
Finally, an end-to-end framework for joint optimization of MPIP design and recovery model will be focused on presenting a more general engine of minimalist and high-quality panoramic imaging.

%% file: Tex_content/appendix_revised.tex
\clearpage
\appendices
\counterwithin{figure}{section}
\counterwithin{equation}{section}
\counterwithin{table}{section}

\section{Sample Images of PALHQ}
\begin{figure}[h]
   \centering
   \includegraphics[width=1.0\linewidth]{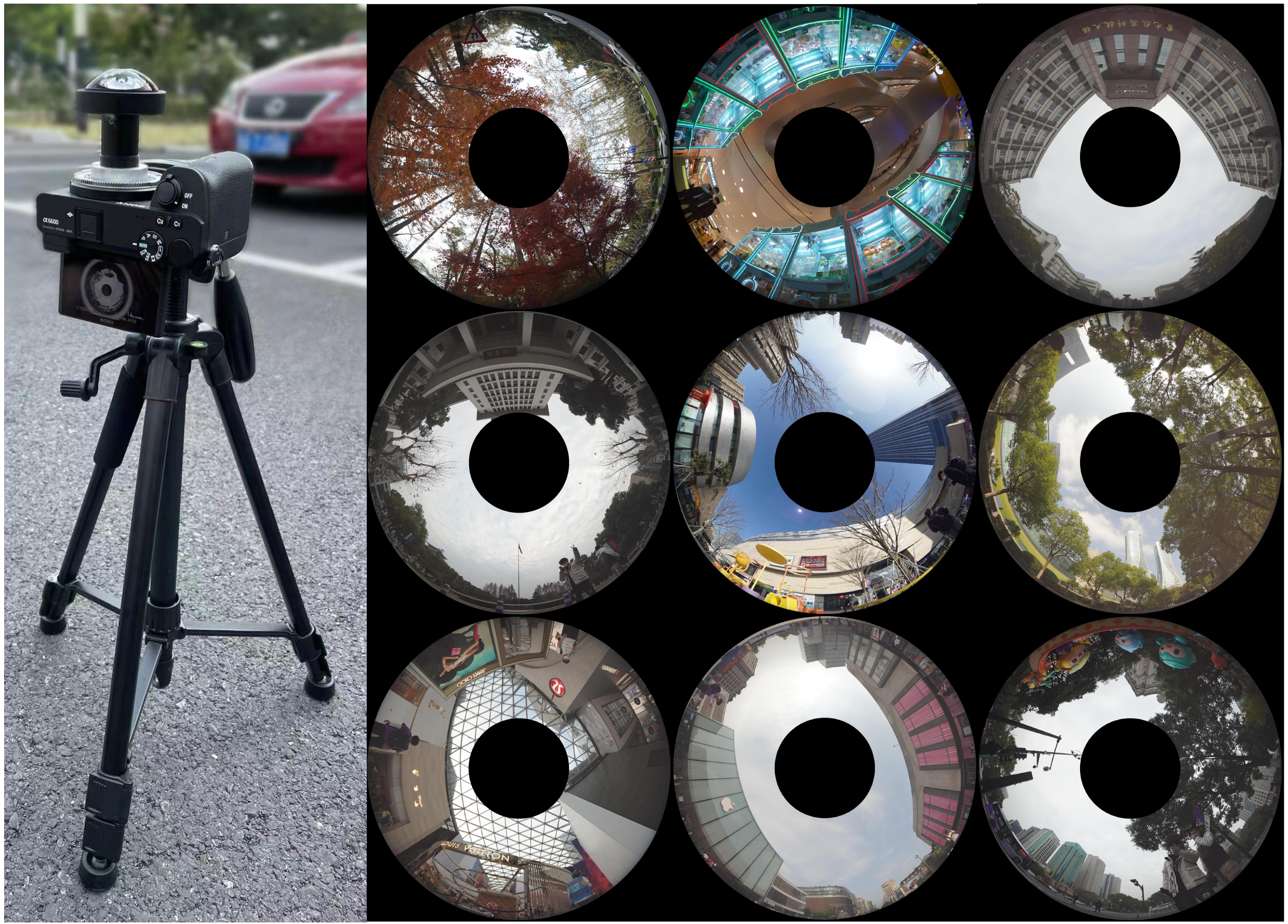}

   \caption{The shooting device and sample images of PALHQ. With a well-designed PAL of $11$ lenses and a Sony $\alpha$6600 camera (on the left), we capture high-quality panoramic images covering various scenes including indoor, natural, urban, campus, and scenic spots (on the right).}
   \label{fig:palhq}

\end{figure}

We show the shooting device and sample images of PALHQ in Fig.~\ref{fig:palhq}. 
The high-quality PAL images dataset covers a wide variety of scenes.
The PAL can present 360{\textdegree} imaging of the surroundings, but with a blind area in the center of the image due to the reflective surface in the center FoV. 
As illustrated on the left of Fig.~\ref{fig:palhq}, the PAL is usually placed toward the sky during the application, where the occlusion of the blind area causes little influence on the acquisition of panoramic information. 
PALHQ serves as the cornerstone of our PCIE for training a robust model to process MPIP images.
Additionally, PALHQ can be transmitted to unfolded panoramas via equirectangular projection (ERP) for other various panoramic image processing applications.
More types of minimalist PAL design, \eg, using fewer lenses or applying meta surface, would also benefit from PALHQ for training learning-based recovery models to improve imaging quality.

\section{Visualization of the PSF Map}
\begin{figure}[h]
   \centering
   \includegraphics[width=1.0\linewidth]{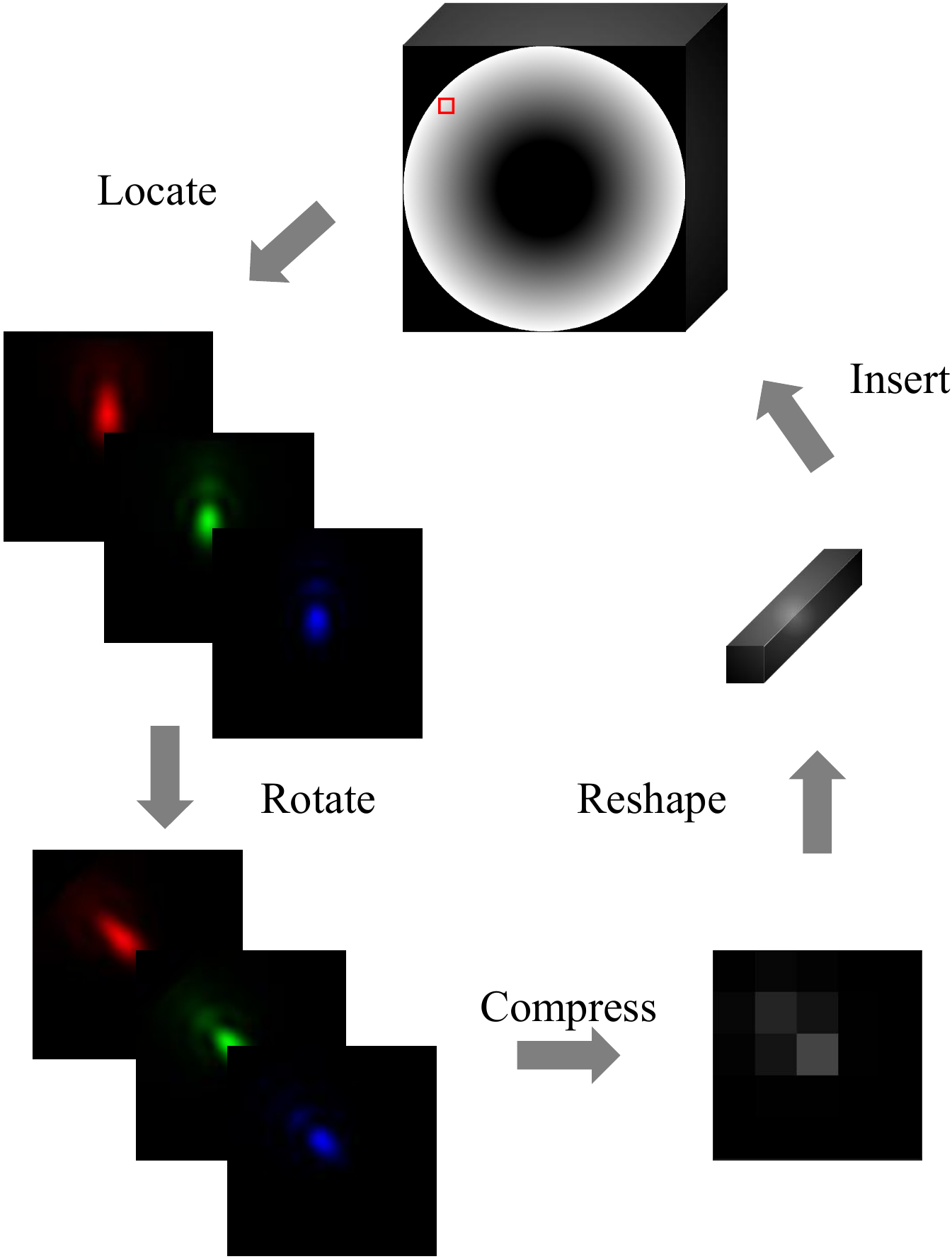}

   \caption{The visualization of compressing PSFs into PSF maps.}
   \label{fig:psfmap}

\end{figure}

The visualization of the process of producing PSF maps is illustrated in Fig.~\ref{fig:psfmap}.
As depicted in Sec.~\YKL{IV-A}, we locate the corresponding FoV of the target pixel and obtain the R/G/B PSFs. 
Then, they are rotated according to their locations and compressed into a shape of ${k'\times{k'}\times1}$ via adaptive average pooling, where $k'$ is set to $5$ as an example.
Finally, we reshape the compressed kernel into channel dimension and insert it into the pixel to produce the PSF map. 

\section{Pipeline for OIQE Calculation}
\begin{figure*}[h]
   \centering
   \includegraphics[width=1.0\linewidth]{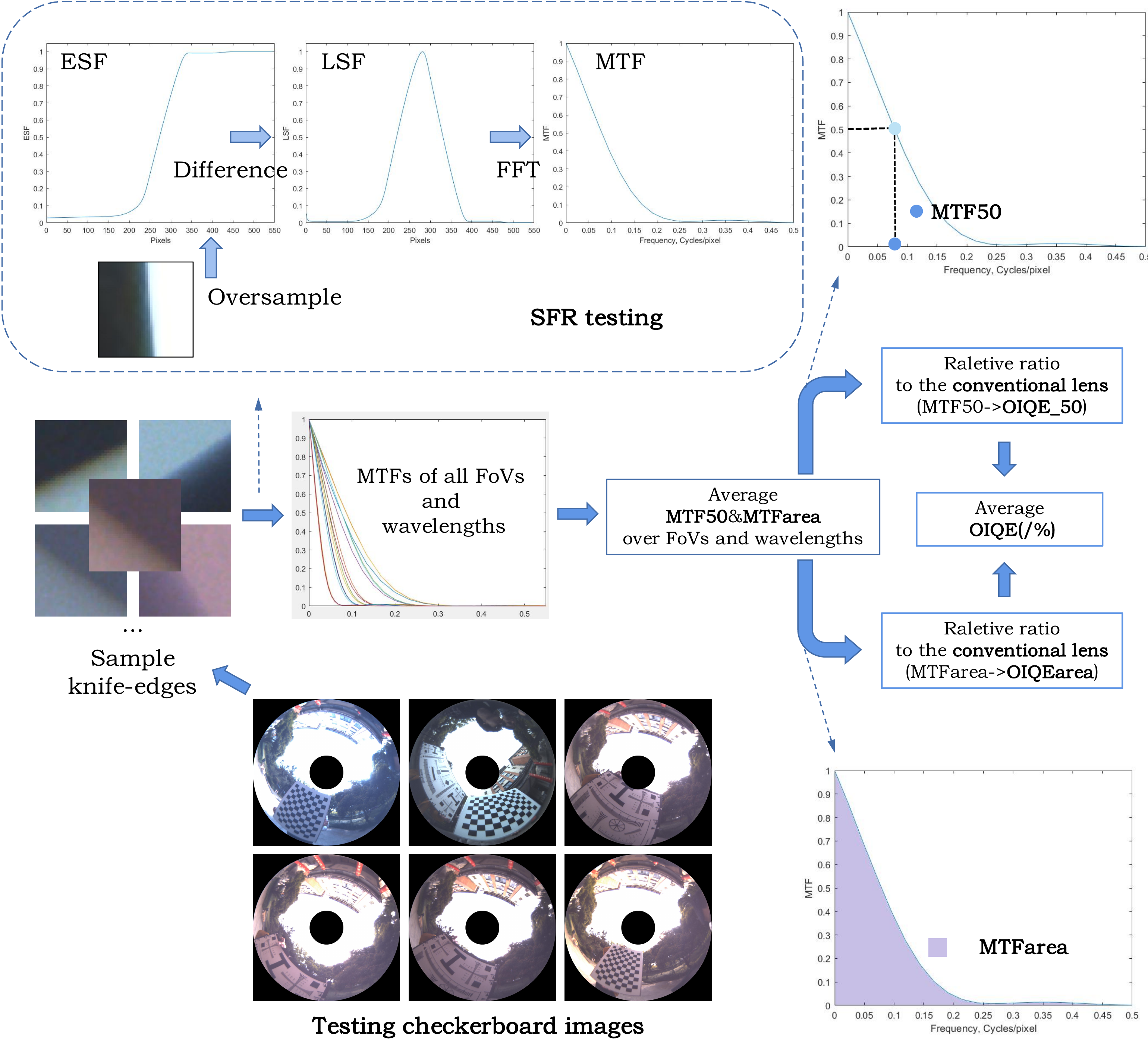}

   \caption{Pipeline for OIQE calculation. ESF: Edge Spread Function. LSF: Line Spread Function.}
   \label{fig:oiqe}

\end{figure*}

The detailed pipeline for calculating the defined OIQE is shown in Fig.~\ref{fig:oiqe}. 
We capture $8$ testing images of the checkerboard under different ISP settings with our MPIP-P1 and AC pipeline. 
For evaluating different models in terms of aberration correction in real-world scenes, the processed testing images are fed into the OIQE pipeline, where sample knife-edges from different FoVs are cropped for SFR testing. 
In OIQE, the comprehensive results of different shooting settings and FoVs present a more credible evaluation of the ability to correct optical aberrations. 

\section{Implementation Details of the Simulation Model}
In this section, we introduce how to generate aberration images based on the simulation model and \textit{Zemax} software in specific. 
To apply Eq.~(\YKL{1}) for simulating aberrations, a set of PSFs under all FoVs of the target optical lens is required. 
We input the structure of MPIP into \textit{Zemax}, then calculate the Zernike standard coefficients under different FoVs ($128$ FoVs from the minimum to the maximum FoV) and wavelengths ($31$ wavelengths from $400{\sim}700nm$), where the first $37$ polynomials are kept as a common practice. 
In this way, the Zernike coefficients matrix with a shape of $31{\times}128{\times}37$ is produced.
Then, we plug the coefficients into Eq.~(\YKL{2}) to describe the wavefronts under all FoVs and wavelengths. 
The random disturbance strategy is applied here to fine-tune the coefficients for multiple virtual aberration distributions.
Before calculating PSFs, we also need to have access to the spot diagram and illumination distribution of the MPIP in \textit{Zemax}, where the sizes of spots determine the kernel sizes (the ratio of the spot size to the pixel size) of PSFs, and the illumination provides the relative amplitude of PSFs. 
Finally, the wavefronts are transformed to PSFs $K_{\theta}(x,y,\lambda)$ via Eq.~(\ref{eq:pupil}) to Eq.~(\ref{eq:intensity}):
\begin{equation}
\label{eq:pupil}
\mathcal{P}_{\theta}(x,y,\lambda) = P(x,y)e^{\mathrm{i}\Phi_{\theta}(x,y,\lambda)},
\end{equation}
\begin{equation}
\label{eq:integrallambda}
E_{\theta}(x,y,\lambda) = \frac{E_0}{{\lambda}d}\iint{\mathcal{P}_{\theta}(x',y',\lambda)e^{-\mathrm{i}\frac{2\pi}{{\lambda}d}(x'x + y'y)}dx'dy'},
\end{equation}
\begin{equation}
\label{eq:intensity}
K_{\theta}(x,y,\lambda) = {|E_{\theta}(x,y,\lambda)|^2},
\end{equation}
where $P(x,y)$ is the circ function and $d$ is the distance from exit pupil to image plane. 
With multiple sets of $K_{\theta}(x,y,\lambda)$ under different random disturbances, we generate aberration images of multiple virtual MPIP samples via Eq.~(\YKL{1}),
where the high-quality MPIP images are transformed to raw images by invert-ISP (Gamma Decompression, Invert Color Correction Matrix, and Invert White Balance), and the aberrated raw images are further processed by ISP (Mosaiced, Adding Noise, Demosaiced, White Balance, Color Correction Matrix, and Gamma Compression) to obtain the final results.

\section{Implementation Details of Model Testing}
During the testing (inference) stage, the input is the full-resolution panoramic images, except for Restormer and NAFNet. 
For global self-attention-based methods, \eg, Restormer, the performance of the model is sensitive to the image resolution, which requires the same resolution during testing and training to maintain consistent high performance~\cite{chu2022improving,beyer2023flexivit}.
Moreover, in our tasks, a larger resolution of testing input represents more complex spatially variant degradation (related to more FoVs), which introduces a larger gap with the training data.
Consequently, in TABLE I of the paper, the results of Restormer are those under the cropping testing strategy, where the input image is cropped into overlapped patches of $256{\times}256$. 
The same is true for NAFNet.
To further illustrate this issue, we test the performance of Restormer and NAFNet under different crop sizes of input, as shown in Table~\ref{tab:resolution}. 
When the training crop size is $256{\times}256$, the performances of the models drop significantly when the testing crop size increases from $256{\times}256$ to $3152{\times}3152$ (the full-resolution). 

\begin{table}[h]
    \begin{center}
        \caption{The impacts of input resolution on Restormer and NAFNet. The models are trained on $256{\times}256$ image patches and tested with different resolutions. We take the results on SynMPIP-P1 as an example. The results in the table are PSNR/SSIM.}

        \label{tab:resolution}
        \input{Table/tab_resolution}
    \end{center}

\end{table}

\section{Ground-Truth Generation Pipeline for Checkerboard Images.}
Capturing real data with Ground Truth (GT) is challenging in the computational imaging field, where no reliable data acquisition pipeline is available in related work.
Taking GT images displayed on the screen with the optical system to be measured could be a solution~\cite{peng2019learned}. However, there still exists a gap between the screen and the real image. At the same time, for a special panoramic system, \ie, MPIP, no suitable screen is available for capturing paired panoramic images.

\begin{figure}
    \centering
    \includegraphics[width=1.0\linewidth]{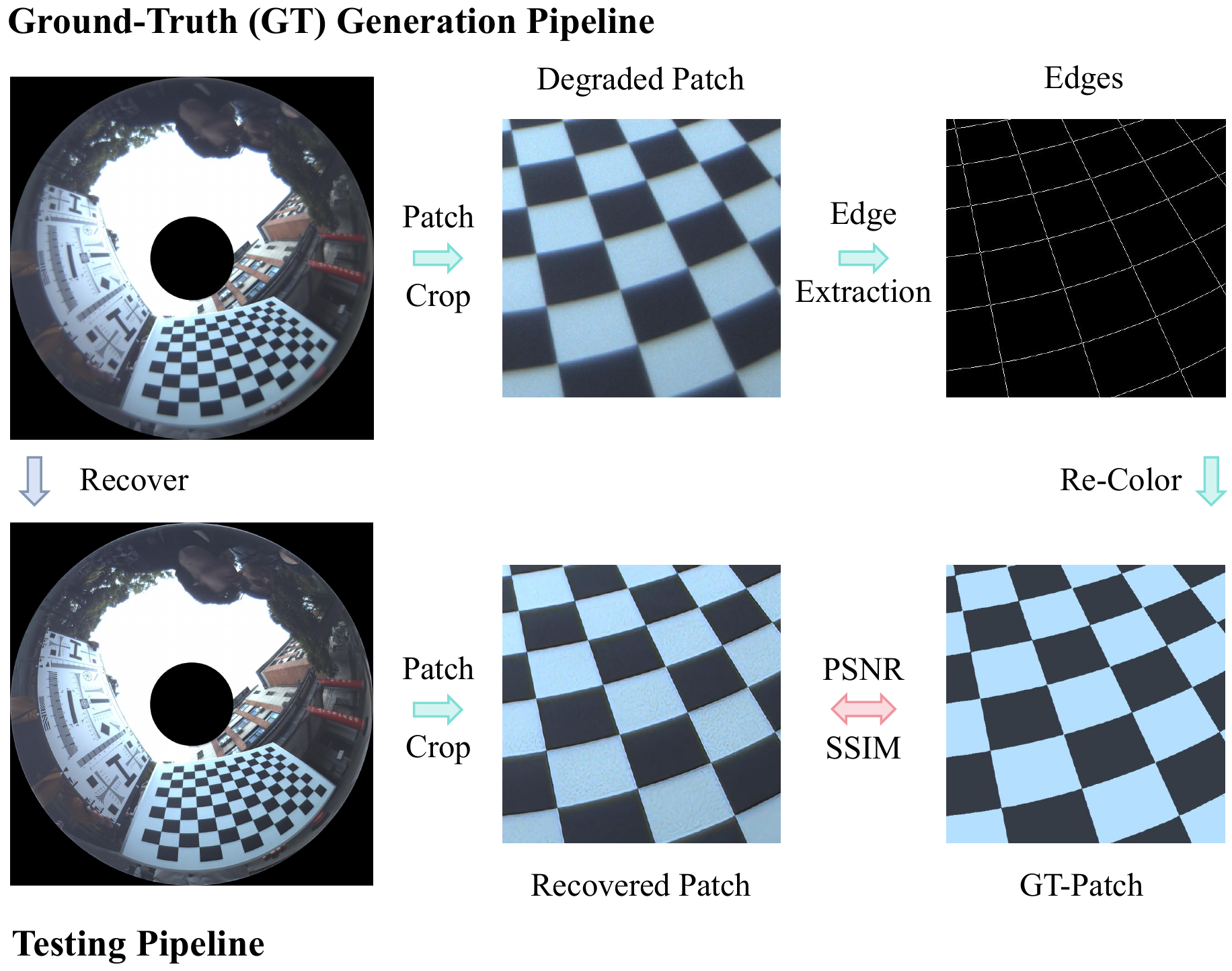}
    \caption{The illustration of the Ground-Truth (GT) generation and quantitative testing pipelines for testing checkerboard images. With the generated GT checkerboard images, we can calculate the error metrics, \eg, PSNR and SSIM, on real-world data. }
    \label{fig:checkertest}
\end{figure}

Consequently, we make an early effort to generate GT images based on captured special patterns. 
For the black and white geometric pattern, \eg, the checkerboard, degraded by aberration degradation, we only need to extract its edge and re-color each part according to its original distribution, to generate its GT pattern. This method was once applied in~\cite{chen2021extreme_quality} to generate checkerboard pairs for training a degradation network. 
In our case, we crop patches of checkerboard test images captured by MPIP, under different FoVs, and generate corresponding GTs by the above method, as shown in Fig.~\ref{fig:checkertest}.
In this way, we only need to crop the patches of the same area on the imaging results of PCIE, and then calculate the error metric, \eg,  PSNR and SSIM, with the corresponding GTs.
The checkerboard testing set of RealMPAL consists of $7$ paired images, where checkerboard patches under different FoVs and ISP settings are included.

However, the pipeline is only a preliminary experiment, which still reveals some weaknesses. For example, the coloring method for GT is worth further investigating, because the chessboard captured by a well-designed PAL is also not as ideal as the GT. The calculation of PSNR and SSIM between recovered images and GT might not fully reflect the ability of aberration correction of the model. Compared to it, the QIQE defined based on the optical metric MTF, is more credible and suitable for evaluating the aberration correction task.

\section{Implementation Details of User Study}
To conduct a subjective evaluation of imaging results of PCIE in real-world scenes, we randomly sample $10$ images from RealMPAL3K and $10$ images from RealMPAL1K for the AC and SR$\&$AC pipeline, respectively. $42$ volunteers are invited to participate in the survey, where they need to go through the imaging results of all the methods in Table~III and select half of them with the best image quality. The final statistical result is presented as the percentage of each method that is being selected, which is the U.S. in Table~III. 

\section{Failure Case of PSF-aware Mechanisms}
From the quantitative results on synthetic datasets, RRDB+ delivers better results than RRDB on most metrics and tasks. However, compared to the significant improvements by PSF-aware mechanisms to SwinIR and GRL, the improvements to RRDB are limited, which even leads to worse FID in some cases (AC-SynMPIP-P1 and SR$\&$AC-SynMPIP-P1).
The OIQE from $66.94\%{\sim}58.28\%$ also illustrates the limitations of PFM on RRDB. 
This is a failure case of the PSF-aware mechanisms.

We speculate that this is caused by the unsatisfactory robustness of the CNN-based model to the domain shift of testing data~\cite{jiang2022computational,kamann2020benchmarking, hoyer2022daformer}. In our evaluation, for both synthetic and real data, the aberration distribution, \ie, the PSF distribution, is slightly different from the standard distribution in PSF representation, which is often the case in the real-world scene due to the manufacture and assembly errors of the lens. In this case, the model has to learn the actual distribution from the standard PSF distribution to guide the PFM. 
Consequently, the CNN-based model is seriously affected by the domain gap, leading to the unreliable prediction of dynamic kernels in PFM, which brings worse performance.

\section{More Results of Generative Models.}
Except for the GAN-based training strategies, the recently developed diffusion model~\cite{ho2020denoising,graikos2022diffusion, jaiswal2023physics} shows strong abilities to generate realistic images with rich details. 
Consequently, we further explore the potential of the diffusion model in our AC task.

\begin{table}[h]
    \begin{center}
        \caption{The quantitative results and computational overhead of SwinIR and corresponding generative models. The denoising steps of SR3 are set to $10$ to make sure that the model can converge. The parameters and FLOPs of SR3 are multiplied by $10$ considering the steps.} 

        \label{tab:ddpm}
        \input{Table/table_ddpm}
    \end{center}

\end{table}

\begin{figure}
    \centering
    \includegraphics[width=1.0\linewidth]{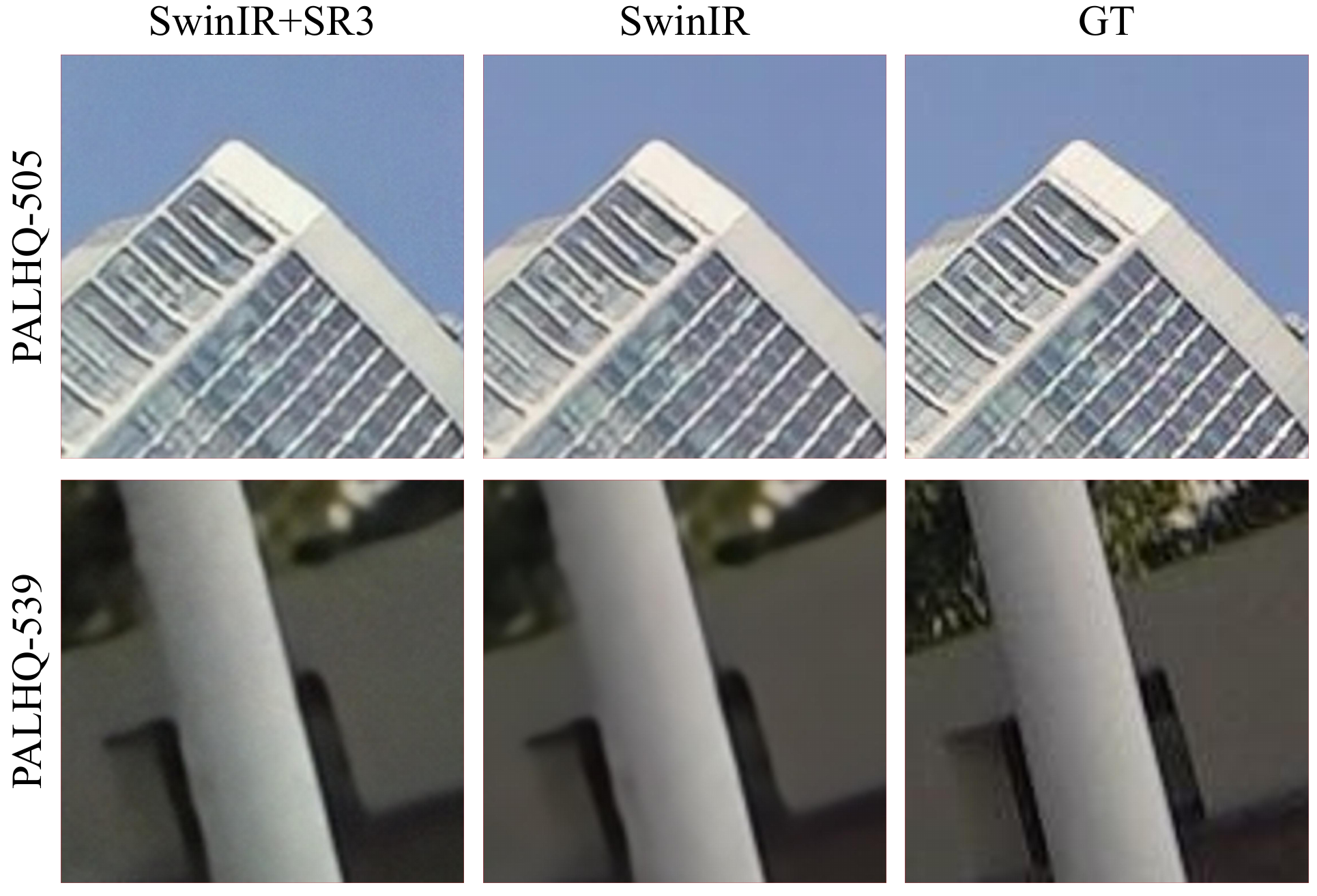}
    \caption{The qualitative results of the refined SwinIR model with SR3. Due to the small denoising steps in our case, the results of SR3 suffer from residual noise and color deviation.}
    \label{fig:ddpm}
\end{figure}

Following~\cite{jaiswal2023physics}, we refine the PSNR-oriented SwinIR model with a diffusion model SR3~\cite{saharia2022image}, where the recovered images of SwinIR are applied as the condition of the diffusion model, considering that the amount of PALHQ is too small to train a diffusion model from scratch. 
The experimental results of the refined model are shown in Fig.~\ref{fig:ddpm} and Table~\ref{tab:ddpm}, where the diffusion model (SwinIR+SR3) cannot bring improvements on perceptual-based metrics like GAN and LDL, but leads to worse performance.

It is known that diffusion models require training on large amounts of datasets under large denoising steps ($2000$ as common practice) for good performance~\cite{ho2020denoising, saharia2022image}.
When the dataset size is small, the number of denoising steps should be set as small to make sure that the model can converge during training ($10$ steps in our case). 
However, the small steps mean a weak denoising ability, leading to the terrible performance of the diffusion model (the residual noise and color deviation).
Moreover, for aberration correction of high-resolution panoramic images, the computational overhead is considerable, where the additional steps of denoiser are unacceptable.
The computational overhead of the SwinIR and additional diffusion model is also shown in Table~\ref{tab:ddpm}, where the Floating Point Operations (FLOPs) are calculated with the input resolution of $1024{\times}1024$. 

In summary, the diffusion model is not suitable for the PCIE currently, but it could be a competitive solution in the future if the PALHQ is developed for larger datasets and an efficient inference pipeline is proposed.

\section{Training Details.}
\begin{figure}
    \centering
    \includegraphics[width=1.0\linewidth]{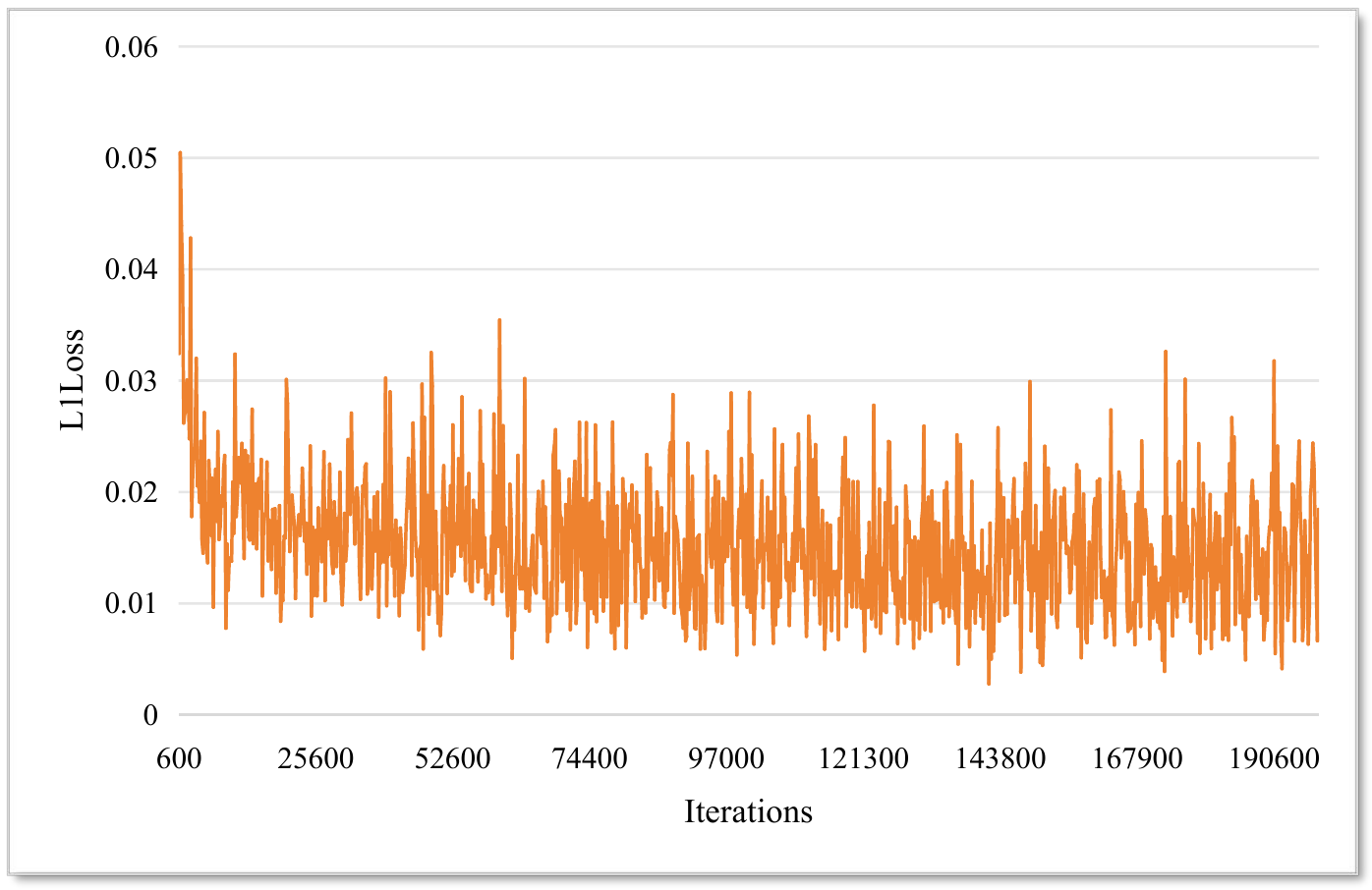}
    \caption{The training loss curve of the PART.}
    \label{fig:loss}
\end{figure}

The parameters of the proposed PART are $19.27M$, which takes $52$ hours for training $200k$ iterations with a batch size of $8$ on a single A800 GPU. Due to the dynamic convolution operations in the model, the small amount of training data, and the guidance of PSF prior, the PSF-aware Transformer can be trained well in small numbers of iterations. The training curve in Fig.~\ref{fig:loss} shows that the model has converged at the end of training. 

\section{The Analysis of Computational Overhead.}

\begin{table*}[t]
    \begin{center}
        \caption{The computational overheads of representative methods. The FLOPs, memory cost, and inference latency are calculated with the input resolution of $1024{\times}1024$ on a single A800 GPU.}

        \label{tab:speed}
        \input{Table/tab_speed}
    \end{center}

\end{table*}

The parameters, FLOPs (Floating Point Operations), memory cost, and inference latency of PSF-aware methods with their baselines are presented in Table~\ref{tab:speed}. The FLOPs, memory cost, and inference latency are calculated with the input resolution of $1024{\times}1024$ on a single A800 GPU, which only needs to be roughly scaled for other input resolutions.

As shown in Table~\ref{tab:speed}, the PSF-aware mechanisms only introduce negligible additional computational overheads ($0.25\%{\sim}1.06\%$ of FLOPs), while bringing significant improvements over the baselines.  
The increase in the number of parameters is mainly due to the the prediction of dynamic convolution kernel of each pixel, which introduces little computational overheads.
The defect of the latency is caused by the transformer-based backbones, while the additional latency brought by PSF-aware mechanisms is not evident ($0.04{\sim}0.2s$).
Benefiting from the efficiency and effectiveness of PSF-aware mechanisms, our future work will focus on more efficient backbones, \eg, lightweight SR backbones~\cite{zhang2022efficient}) to achieve light-weight and high-quality panoramic imaging.

Moreover, we also release a lightweight version of PART, \ie, PART-S (with smaller depth and embedding dim), considering the potential applications of the PCIE in mobile and wearable terminals. 
With only $19.00\%$ of the computational overhead, PART-S can achieve comparable performance to the baseline SwinIR.

%% file: Table/tab_resolution.tex
\resizebox{0.4\textwidth}{!}
{
\renewcommand{\arraystretch}{1.2}
\setlength{\tabcolsep}{1mm}{

\begin{tabular}{c|ccc}
\hline
\multirow{2}{*}{Method} & \multicolumn{3}{c}{Input Resolution} \\ \cline{2-4} 
 & 3152 & 1024 & 256 \\ \hline\hline
Restormer &27.424/0.8826  &30.333/0.9088  &32.971/0.9287  \\ 
NAFNet &28.494/0.8853  &30.481/0.9099  &32.837/0.9274  \\ \hline
\end{tabular}

}
}

%% file: Table/table_ddpm.tex
\resizebox{0.50\textwidth}{!}
{
\renewcommand{\arraystretch}{1.2}
\setlength{\tabcolsep}{1mm}{

\begin{tabular}{c|cccc|cc}
\hline
 & PSNR & SSIM & LPIPS & FID & Params & FLOPs \\ \hline\hline
SwinIR & 32.913 & 0.9291 & 0.0446 & 03.670 & 11.97M &407.76G  \\
SwinIR+GAN & 29.916 & 0.8920 & 0.0449 & 04.254 & 11.97M &407.76G  \\
SwinIR+LDL & 31.770 & 0.9130 & 0.0297 & 03.444 & 11.97M &407.76G  \\
SwinIR+SR3 & 31.281 & 0.8412 & 0.4110 & 17.956 & 11.97M+13.85M &407.76G+3481G  \\ \hline
\end{tabular}

}
}

%% file: Table/tab_speed.tex
\resizebox{0.7\textwidth}{!}
{
\renewcommand{\arraystretch}{1.2}
\setlength{\tabcolsep}{1mm}{
\begin{tabular}{cc|cccc|cccc}
\hline
\multicolumn{2}{c|}{Method} & PSNR & SSIM & LPIPS & FID & Params & FLOPs & Memory & Latency \\ \hline\hline
\multicolumn{1}{c|}{\multirow{3}{*}{Baselines}} & RRDB & 32.716 & 0.9265 & 0.0469 & 03.704 & 16.72M & 588.28G &0.58GB  &0.15s  \\
\multicolumn{1}{c|}{} & SwinIR & 32.913 & 0.9291 & 0.0446 & 03.670 & 11.97M & 407.76G &0.65GB  &0.97s  \\
\multicolumn{1}{c|}{} & GRL & 32.369 & 0.9256 & 0.0457 & 04.234 & 20.27M & 649.24G &1.23GB  &1.37s  \\ \hline
\multicolumn{1}{c|}{\multirow{4}{*}{PSF-aware}} & RRDB+ & 32.816 & 0.9271 & 0.0456 & 03.746 & 18.61M & 589.75G &0.83GB  &0.27s  \\
\multicolumn{1}{c|}{} & GRL+ & 32.847 & 0.9292 & 0.0454 & 03.627 & 25.60M & 653.43G &2.11GB  &1.41s  \\
\multicolumn{1}{c|}{} & PART & 33.143 & 0.9304 & 0.0435 & 03.571 & 19.27M & 412.08G &1.90GB  &1.17s  \\
\multicolumn{1}{c|}{} & PART-S & 32.812 & 0.9278 & 0.0452 & 03.710 & 03.89M & 077.47G &1.00GB  &0.73s  \\ \hline
\end{tabular}

}
}